\documentclass[letterpaper]{article} 
\usepackage[preprint]{aaai2027}  
\usepackage[hyphens]{url}  
\usepackage{graphicx} 
\usepackage{natbib}  
\usepackage{caption} 
\usepackage{amsmath}
\usepackage{amssymb}
\usepackage{booktabs}
\usepackage{multirow}
\usepackage{algorithm}
\usepackage{algorithmic}
\title{ProCon: Projection-Consistency Memory for Training-Free Anomaly Detection}
\author{
JOONGWON CHAE\textsuperscript{\rm *1,2},
Lihui Luo\textsuperscript{\rm *1},
Yang Liu\textsuperscript{\rm 1},
Dongmei Yu\textsuperscript{\rm 3},
Peiwu Qin\textsuperscript{\rm 4},
Runming Wang\textsuperscript{\rm 1},
ILMOON CHAE\textsuperscript{\rm \dag 2}
}
\affiliations{
\textsuperscript{\rm 1}Tsinghua University Shenzhen International Graduate School, Shenzhen, China\\
\textsuperscript{\rm 2}Ratel Soft\\
\textsuperscript{\rm 3}Affiliated Fifth Hospital, Wenzhou Medical University, Wenzhou, Zhejiang, China\\
\textsuperscript{\rm 4}Chinese Medicine Guangdong Laboratory\\
\textsuperscript{\rm *}These authors contributed equally.\quad\textsuperscript{\rm \dag}Corresponding author.
}

\begin{document}

\maketitle

\begin{abstract}
Memory-based anomaly detection is attractive because it localizes defects from normal images without training a decoder or synthesizing pseudo anomalies. However, most memory methods still use the memory bank as a nearest-neighbor lookup table: a test patch is treated as normal if it has one nearby normal anchor. This hard retrieval view is vulnerable to false-normal matches and does not test whether the patch is consistently supported by a local normal neighborhood. We propose ProCon, a training-free framework that turns memory retrieval into decoder-free reconstruction. ProCon softly projects each test patch onto nearby normal memory vectors and uses the projection residual as anomaly evidence. To stabilize this residual, it constructs seed-perturbed layer-wise memories, aggregates bank residuals by a median, and fuses depth-specific residual maps by layer consensus. ProCon requires no decoder training, backbone fine-tuning, learned fusion weights, or pseudo-anomaly supervision. Across MVTec-AD, VisA, and Real-IAD under the single-category evaluation protocol, ProCon achieves strong image- and pixel-level performance under seven standard metrics, including image AUROC scores of 99.8\%, 99.2\%, and 93.2\%, respectively. Ablations show that the gains come from replacing hard retrieval with soft normal projection and stabilizing the residuals through memory and depth consensus.
The code is available at \url{https://github.com/jw-chae/Procon}.

\end{abstract}

\section{Introduction}

Industrial anomaly detection aims to identify and localize defects using only normal training images. This normal-only setting is important for visual inspection, where anomalous samples are rare, incomplete, or expensive to collect. A practical detector must therefore output both an image-level abnormality score and a pixel-level anomaly map. Benchmarks such as MVTec-AD~\cite{bergmannMVTecADComprehensive2019}, VisA~\cite{zou2022spot}, and Real-IAD~\cite{wang2024realiad} evaluate these two requirements jointly.

Memory-based methods are strong in this setting because they reuse pretrained features and avoid learning a decoder. SPADE, PaDiM, and PatchCore compare test features with stored normal references, often through nearest-neighbor retrieval or spatial distribution matching~\cite{cohen2020spade,defard2021padim,roth2022patchcore}. The dominant PatchCore-style score for a patch feature $z$ is
\begin{equation}
 s_{\mathrm{NN}}(z)=\min_{m\in M}\|z-m\|_2,
\end{equation}
where $M$ is a normal memory bank. This score can be viewed as a hard projection onto one normal anchor. It asks whether there exists a nearby normal reference, but not whether $z$ can be explained by the local normal neighborhood around that reference.

This distinction matters. A local defect can accidentally match one normal anchor in a high-dimensional pretrained feature space, especially when the feature is semantically abstract or the memory covers a broad range of appearances. Such a false-normal match produces a small nearest-neighbor distance even though the patch is not consistently supported by the normal manifold. Our central thesis is that memory-based anomaly detection should not only ask whether a test patch is close to one normal anchor; it should ask whether the patch is locally reconstructible from normal memory.

We propose \textbf{ProCon} (\textbf{Pro}jection \textbf{Con}sistency), a training-free method that uses normal memory as a non-parametric projection operator. For each test patch, ProCon retrieves nearby normal anchors, computes distance-based soft weights, projects the patch onto their weighted normal estimate, and scores the projection residual. This converts memory retrieval into decoder-free reconstruction: no decoder is trained, but the residual still measures how well normal references can explain the patch.

ProCon further stabilizes this residual with two consensus axes. First, seed-perturbed coreset banks are built for each layer, and their residual maps are aggregated by a median to reduce sensitivity to unlucky memory samples. Second, independent memories are built for selected DINOv2 depths, and layer-wise residual maps are averaged only after each layer has produced the same soft-projection residual quantity. This avoids collapsing layer-specific normal geometry by premature feature concatenation.

Our contributions are threefold. First, we reinterpret nearest-neighbor memory scoring as hard projection and replace it with soft local normal projection. Second, we introduce a projection-consistency memory framework with bank-wise and layer-wise consensus, while remaining decoder-free and training-free. Third, we validate ProCon on MVTec-AD, VisA, and Real-IAD with seven image- and pixel-level metrics, and show through controlled ablations that the improvement is due to projection residuals and consensus rather than simply larger memories.

\begin{figure*}[t]
\centering
\includegraphics[width=0.98\textwidth]{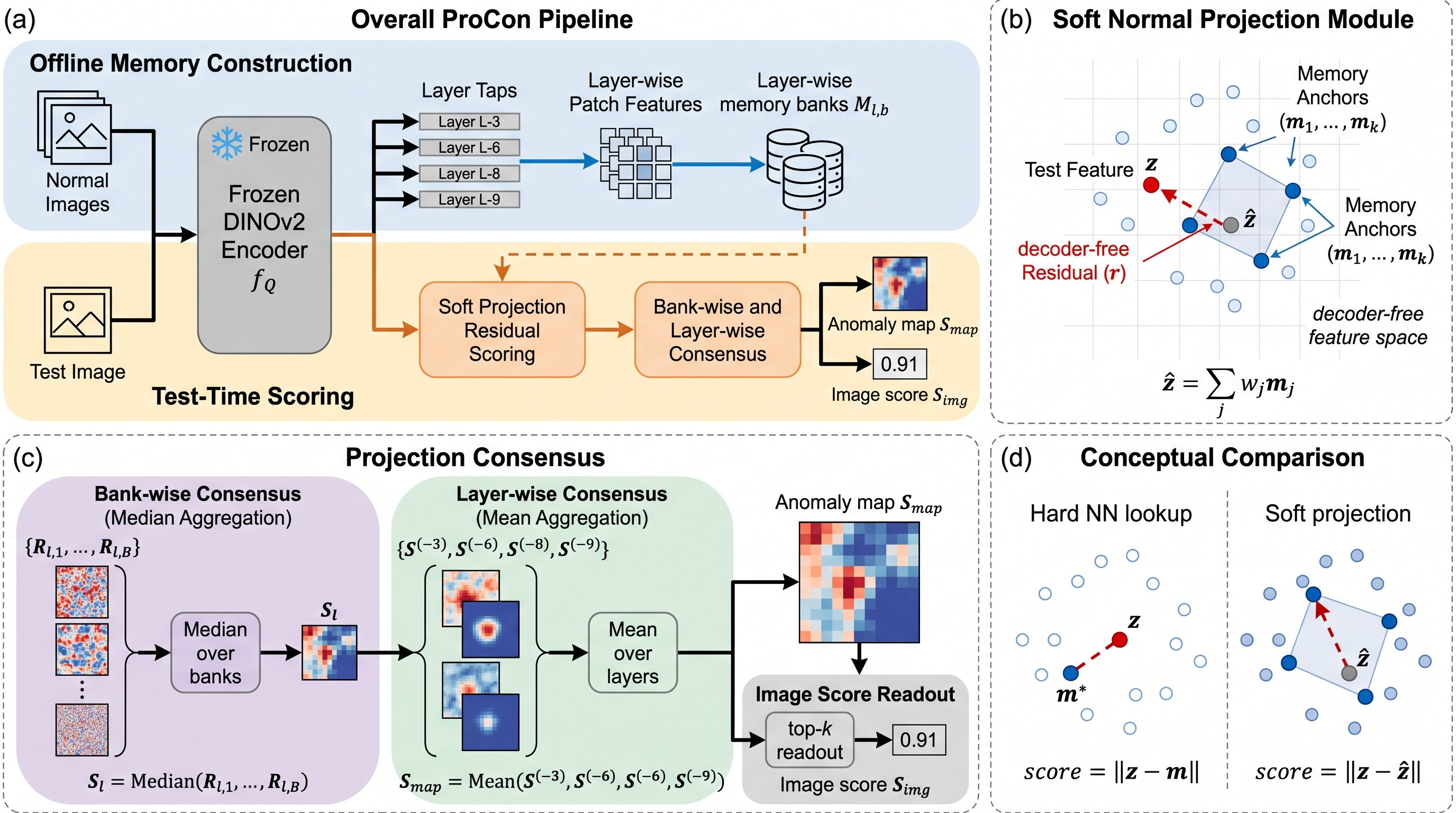}
\caption{Overview of ProCon. Normal images are encoded by a frozen DINOv2 backbone, and independent layer-wise coreset memories are built for selected depths. At test time, each patch is softly projected onto nearby normal anchors. The residual maps are aggregated by bank-wise median consensus and layer-wise mean consensus to produce the final anomaly map and image score.}
\label{fig:overview}
\end{figure*}

\section{Related Work}

\subsection{Memory-based Anomaly Detection and Memory Redesign}

Memory-based anomaly detection detects defects by comparing test features with normal references stored from anomaly-free training images. Early feature-retrieval methods such as SPADE use pretrained representations and nearest-neighbor correspondences between test images and normal samples for anomaly localization~\cite{cohen2020spade}. PaDiM models normal patch embeddings with spatially aligned feature distributions~\cite{defard2021padim}. PatchCore establishes a strong memory-bank paradigm by storing a representative coreset of nominal patch features and scoring test patches by their nearest-neighbor distance to this memory~\cite{roth2022patchcore}. These methods are attractive because they provide strong localization from pretrained features without decoder training or pseudo-anomaly synthesis.

Following PatchCore, several studies have shown that memory design is a central factor in industrial anomaly detection. SoftPatch improves robustness under noisy normal training data by stabilizing patch-level memory construction~\cite{jiang2022softpatch}. CFA and ReConPatch reduce the mismatch between generic pretrained features and target industrial domains through feature adaptation or contrastive patch representation learning~\cite{lee2022cfa,hyun2024reconpatch}. GraphCore and Divide-and-Assemble further study memory redundancy, invariant features, and block-wise memory modules~\cite{xie2023graphcore,hou2021divideassemble}. Recent extensions
have broadened the role of memory-bank models beyond plain nearest-neighbor
scoring. StructCore shows that frozen DINO-based patch memories equipped with
structure-aware image-level scoring can match learned Transformer reconstruction
methods without any training~\cite{chae2026structcore}, and GCR extends coreset
memories to continual multi-category settings, attributing performance collapse
to cross-task routing instability rather than representation forgetting and
addressing it with geometry-consistent routing~\cite{chae2026gcr}.

ProCon follows this memory-redesign line, but changes the use of memory from nearest-neighbor retrieval to soft normal projection: the memory bank is used not as a reference table but as a non-parametric reconstruction operator whose projection residual becomes the anomaly evidence.

\subsection{Reconstruction-based and Prototype-based Anomaly Detection}

Another major family obtains anomaly evidence from reconstruction errors, teacher-student discrepancies, likelihood estimation, or learned normal prototypes. RD4AD reconstructs pretrained teacher features and uses feature residuals as anomaly scores~\cite{deng2022rd4ad}. DRAEM learns a reconstruction-discrimination framework with synthetic anomalies~\cite{zavrtanik2021draem}, while UniAD studies unified reconstruction for multi-class anomaly detection and analyzes shortcut behavior in which anomalies may also be reconstructed as normal~\cite{you2022uniad}. Flow-based methods such as FastFlow and CFLOW-AD learn likelihood models over pretrained feature maps~\cite{yu2021fastflow,gudovskiy2022cflow}, and other high-performing methods use denoising student-teacher learning, synthetic feature perturbation, discriminative scoring, or efficient teacher-student designs~\cite{zhang2023destseg,liu2023simplenet,batzner2024efficientad}.

Recent Transformer-based reconstruction and prototype methods have further advanced the field. Dinomaly proposes a minimalist Transformer reconstruction framework for multi-class unsupervised anomaly detection~\cite{guo2025dinomaly}, and INP-Former extracts image-conditioned normal prototypes from the intrinsic normal regions of a test image to guide reconstruction~\cite{luo2025inpformer}. These methods demonstrate the strength of reconstruction-based normality modeling, but rely on learned reconstruction modules, prototype extractors, or training objectives. ProCon also produces residual-style evidence, but computes it non-parametrically from frozen memory: it connects memory-based retrieval and reconstruction-based scoring by turning normal memory into a decoder-free reconstruction mechanism.

\subsection{Consensus, Ensembles, and Multi-reference Scoring}

Classical ensemble studies show that combining predictors improves generalization when the members are individually reliable and make partially different errors~\cite{dietterich2000ensemble,hansen1990neural,krogh1994neural,tumer1996error}, whether the diversity comes from perturbed samples~\cite{breiman1996bagging} or from different feature subspaces~\cite{ho1998random}; similar ideas appear in outlier-detection ensembles~\cite{lazarevic2005feature,zimek2013ensembles,rayana2016less,zhao2019lscp}. ProCon adopts this perspective in a training-free memory setting: seed-perturbed coreset banks supply memory diversity along one axis, independent depth-specific memories supply representation diversity along the other, and---unlike generic ensemble fusion---aggregation happens only after every bank and layer has been converted into the same soft-projection residual form.

\section{Method}

\subsection{Overview}

We propose \textbf{ProCon} (\textbf{Pro}jection \textbf{Con}sistency), a
training-free anomaly detection framework that turns memory retrieval into
decoder-free reconstruction. ProCon uses a frozen DINOv2 ViT-B/14
backbone~\cite{oquab2023dinov2,dosovitskiy2021vit} and never trains a decoder,
fine-tunes the backbone, learns fusion weights, or uses pseudo-anomaly
supervision.

The method consists of two stages. In the memory construction stage, patch
features of normal training images are extracted from a selected set of
transformer layers, and independent normal coreset memories are built for each
layer rather than for a concatenation of all layers. In the inference stage,
each test patch is softly projected onto nearby normal memory vectors, and the
residual between the feature and its soft normal projection is the anomaly
evidence. ProCon stabilizes this residual through two consensus axes: a
bank-wise median over seed-perturbed coreset memories, and a layer-wise mean
over depth-specific residual maps.

The central design principle is that fusion is performed \emph{after} residual
alignment, not before memory construction. Every bank and every selected layer
first produces the same quantity---a soft-projection residual with respect to a
local normal memory neighborhood---and only these aligned residuals are
aggregated.

\subsection{Problem Setup and Notation}

Let $\mathcal{D}_{\mathrm{train}}=\{x_i\}_{i=1}^{N}$ be a set of anomaly-free
training images. Given a test image $x$, the goal is to produce a pixel-level
anomaly map $S_{\mathrm{map}}$ and an image-level anomaly score
$S_{\mathrm{img}}$.

We use a frozen feature extractor $f_\theta$ and extract patch tokens from a set
of selected transformer layers
\[
\mathcal{L}=\{-3,-6,-8,-9\}.
\]
Negative indices count backward from the last transformer block, so that $-3$ is
the third layer from the top and $-9$ is a shallower, more localization-oriented
layer. All blocks emit the same patch grid, so layer maps are spatially aligned
with no resolution confound. For an image $x$ and layer $\ell\in\mathcal{L}$,
the patch feature map is
\[
F_\ell(x)=f_\theta^\ell(x)\in\mathbb{R}^{P\times C},
\]
where $P$ is the number of patch tokens and $C$ is the feature dimension. We
denote the $p$-th patch feature at layer $\ell$ by
$z_{\ell,p}\in\mathbb{R}^{C}$.

For each layer $\ell$, ProCon constructs $B$ independent memory banks
$\mathcal{M}_{\ell}=\{M_{\ell,1},\dots,M_{\ell,B}\}$, where each $M_{\ell,b}$ is
a coreset selected from the normal patch features of layer $\ell$ under a
seed-perturbed construction. We use $B=5$. The coreset ratio is denoted by
$\rho_m$; in our final configuration $\rho_m=5\%$ for MVTec-AD and VisA, and
$\rho_m=1\%$ for Real-IAD.

\subsection{Nearest-neighbor Memory as Hard Projection}

PatchCore-style memory methods score a test patch by its distance to the nearest
normal memory vector~\cite{roth2022patchcore}. Given a memory bank $M$, this can
be interpreted as a \emph{hard projection} onto a single normal anchor:
\begin{gather*}
\Pi^{\mathrm{hard}}_M(z)=\arg\min_{m\in M}\|z-m\|_2,\\
s_{\mathrm{NN}}(z)=\bigl\|z-\Pi^{\mathrm{hard}}_M(z)\bigr\|_2.
\end{gather*}
This formulation shows that nearest-neighbor retrieval only asks whether there
\emph{exists} one normal anchor close to the test patch. An anomalous patch can
therefore receive a low score if it accidentally matches a single normal memory
vector, even when it is not consistently supported by the local normal
manifold. We refer to this failure mode as a \emph{false-normal match}.

\subsection{Soft Local Normal Projection}

ProCon replaces hard retrieval with soft local normal projection. For a test
patch $z_{\ell,p}$, layer $\ell$, and memory bank $M_{\ell,b}$, we first retrieve
its $k$ nearest normal memory vectors and write
$\mathcal{N}_k=\mathrm{kNN}(z_{\ell,p},M_{\ell,b})$ for this neighborhood. Each
neighbor $m_j\in\mathcal{N}_k$ receives a distance-based soft weight
\[
w_j=
\frac{\exp\!\left(-\|z_{\ell,p}-m_j\|_2^2/\tau_{\ell,b}\right)}
{\sum_{m_t\in\mathcal{N}_k}\exp\!\left(-\|z_{\ell,p}-m_t\|_2^2/\tau_{\ell,b}\right)}.
\]
The soft normal projection and the projection residual are
\[
\hat z_{\ell,b,p}=\sum_{m_j\in\mathcal{N}_k} w_j\, m_j,
\qquad
r_{\ell,b,p}=\bigl\|z_{\ell,p}-\hat z_{\ell,b,p}\bigr\|_2 .
\]
This residual is reconstruction-like, but no reconstruction model is trained:
the normal memory itself acts as a non-parametric reconstruction operator. A
normal patch should be well explained by a local neighborhood of normal
references, whereas an anomalous patch leaves a larger residual. The
distance-based weighting is essential: with uniform weights the projection
collapses to the neighborhood centroid regardless of how the neighbors are
arranged around the test patch, blurring the distinction between a patch inside
the local normal manifold and one merely equidistant from several anchors
outside it.

We use $k=5$. The temperature $\tau_{\ell,b}$ is not a fixed constant but is
computed adaptively \emph{for each test image and each bank},
\[
\tau_{\ell,b}
=\operatorname{median}
\bigl(\{\,\|z_{\ell,p}-m_j\|_2^2 : m_j\in\mathcal{N}_k\,\}_{p,j}\bigr),
\]
where $p$ ranges over the patches of the current test image. Tying the softmax
bandwidth to the bank's local normal spacing keeps the residual discriminative
across categories, depths, and memory samples.

\subsection{Bank-wise Projection Consensus}

A single coreset memory can be sensitive to the particular normal anchors
selected during coreset construction. To reduce this sampling sensitivity,
ProCon builds $B$ seed-perturbed coreset banks for each layer. Each bank
produces its own projection-residual map $R_{\ell,b}(p)=r_{\ell,b,p}$, and the
bank-wise consensus for layer $\ell$ is the median residual over banks,
\[
S_{\ell}(p)
=\operatorname*{median}_{b=1,\dots,B} R_{\ell,b}(p).
\]
The median suppresses unlucky memory samples that would otherwise produce a
false-normal match for some anomalous patch. This is the first consensus axis of
ProCon: projection consensus over memory perturbations.

\subsection{Layer-wise Independent Normal Memories}

Multi-layer memory methods often concatenate features from several layers
\emph{before} memory construction,
\[
M_{\mathrm{concat}}
=\mathrm{Coreset}\!\left(
[\,F_{\ell_1};F_{\ell_2};\dots;F_{\ell_n}\,]
\right).
\]
Although concatenation increases feature dimensionality, it collapses
layer-specific normal geometries into one mixed coreset space. The selected
anchors may then be dominated by the layers with larger variance, weakening the
contribution of layers that are useful for fine localization.

ProCon avoids this collapse by constructing independent memories for each
selected layer,
\[
M_{\ell,b}
=\mathrm{Coreset}\!\left(
G_\ell\bigl(F_\ell(\mathcal{D}_{\mathrm{train}})\bigr);\, b,\rho_m
\right)
\]
for each $\ell\in\mathcal{L}$ and $b=1,\dots,B$, where $G_\ell$ is a fixed,
layer-specific transformation applied before coreset selection and test-time
scoring. In practice, $G_\ell$ applies feature normalization followed by a fixed
random projection to a lower-dimensional space for efficient memory construction
and retrieval; no projection weights are learned. Full construction details are
provided in the supplementary material.

This design preserves depth-specific normal geometry. Each layer independently
converts its normal memory into a projection-residual map $S_\ell$, and layer
fusion is performed only after this residual alignment: the residuals carry the
same semantic meaning across layers---unexplained distance from a local normal
projection---before they are combined.

\subsection{Depth-wise Projection Consensus and Readout}

Given the bank-consensed residual maps $\{S_\ell\}_{\ell\in\mathcal{L}}$, the
final patch-level anomaly map is a fixed mean over the selected depths,
\[
S_{\mathrm{map}}(p)
=\frac{1}{|\mathcal{L}|}
\sum_{\ell\in\mathcal{L}} S_\ell(p).
\]
We use the raw residual scale and do not apply per-image normalization before
layer fusion; pixel-level localization is read directly from $S_{\mathrm{map}}$.
This is the second consensus axis of ProCon: projection consensus over
representation depth.

For image-level detection, ProCon converts the anomaly map into an image score
by top-mean pooling. Let $\mathcal{T}_{\rho}$ be the set of patch indices
corresponding to the top-$\rho$ fraction of scores in $S_{\mathrm{map}}$. The
image score is
\[
S_{\mathrm{img}}
=\frac{1}{|\mathcal{T}_{\rho}|}
\sum_{p\in\mathcal{T}_{\rho}}
S_{\mathrm{map}}(p),
\]
with $\rho=0.005$. The top-mean readout affects only the image-level score;
pixel-level metrics are computed from the full anomaly map.

\subsection{Complexity}

The only data-dependent operation before inference is normal memory
construction through coreset selection; no parameters are optimized by
back-propagation. At inference, ProCon performs $|\mathcal{L}|\,B$
soft-projection evaluations per test image. If each memory bank stores $|M|$
vectors, each image has $P$ patches, and the projected feature dimension is $D$,
the dominant retrieval cost is $O(|\mathcal{L}|\,B\,P\,|M|\,D)$. In our final
configuration, $|\mathcal{L}|=4$ and $B=5$, giving $20$ layer-bank projection
evaluations per image. All selected layers are extracted from a single frozen
DINOv2 forward pass, and memory banks are stored in the projected feature space
for efficient nearest-neighbor search. The full procedure and pseudocode are
provided in the supplementary material.

\section{Experiments}

\begin{table*}[!t]
\centering
\small
\renewcommand{\arraystretch}{1.07}
\setlength{\tabcolsep}{4.2pt}
\begin{tabular*}{\textwidth}{@{\extracolsep{\fill}}lccc ccc ccc}
\toprule
& \multicolumn{3}{c}{\textbf{MVTec-AD}} & \multicolumn{3}{c}{\textbf{VisA}} & \multicolumn{3}{c}{\textbf{Real-IAD}} \\
\cmidrule(lr){2-4}\cmidrule(lr){5-7}\cmidrule(lr){8-10}
\textbf{Method} & I-AUROC & P-AP & AUPRO & I-AUROC & P-AP & AUPRO & I-AUROC & P-AP & AUPRO \\
\midrule
PatchCore~\cite{roth2022patchcore}   & 99.1 & 56.1 & 93.5 & 95.1 & 40.1 & 91.2 & 89.4 & --   & 91.5 \\
RD4AD~\cite{deng2022rd4ad}           & 98.5 & 58.0 & 93.9 & 96.0 & 27.7 & 70.9 & 87.1 & --   & 93.8 \\
SimpleNet~\cite{liu2023simplenet}    & 99.6 & 54.8 & 90.0 & 96.8 & 36.3 & 88.7 & 88.5 & --   & 84.6 \\
Dinomaly~\cite{guo2025dinomaly}      & 99.7 & 68.9 & 95.0 & 98.9 & 50.7 & 95.1 & 92.0 & 45.2 & 95.1 \\
INP-Former~\cite{luo2025inpformer}   & 99.7 & 70.2 & 95.4 & 98.5 & 49.2 & 93.8 & 92.1 & 48.1 & 95.6 \\
\textbf{ProCon}                      & \textbf{99.8} & \textbf{73.5} & \textbf{95.9} & \textbf{99.2} & \textbf{52.3} & \textbf{97.0} & \textbf{93.2} & \textbf{49.4} & \textbf{97.2} \\
\bottomrule
\end{tabular*}
\caption{Main single-category comparison using the common published metrics of
recent single-category Dinomaly and INP-Former evaluations. Baseline numbers are
from the original papers under the single-category protocol. ProCon uses a $5\%$
coreset on MVTec-AD and VisA and a $1\%$ coreset on Real-IAD (see text). All
values are percentages.}
\label{tab:main_comparison}
\end{table*}

\subsection{Experimental Setup}

We evaluate ProCon on MVTec-AD, VisA, and Real-IAD. MVTec-AD and VisA are
evaluated in the standard single-category setting, while Real-IAD uses the
single-view protocol with category-specific normal memories. We report three
image-level metrics (I-AUROC, I-AP, and max I-F1), three pixel-level metrics
(P-AUROC, P-AP, and max P-F1), and AUPRO. All numbers are percentages.

The coreset ratio is $5\%$ for MVTec-AD and VisA and $1\%$ for Real-IAD. These
budgets follow from two observations in our budget analysis
(Table~\ref{tab:budget}). First, both ProCon and its single-layer baseline
saturate at $5\%$: the $1\%\!\to\!5\%$ step gives a small consistent gain while
$5\%\!\to\!10\%$ is essentially flat, so $5\%$ is the accuracy-cost sweet spot on
the small benchmarks. Second, Real-IAD contains substantially more normal
training images per category, so a $1\%$ coreset already yields a dense normal
memory. Unless noted otherwise, ProCon uses DINOv2 ViT-B/14, layers
$\{-3,-6,-8,-9\}$, $B=5$ banks per layer, $k=5$ neighbors, bank-wise median
aggregation, layer-wise mean aggregation, and top-mean image readout. We denote
the single-layer soft-projection memory baseline as SPM; it corresponds to the
third row of the ablation ladder (Table~\ref{tab:component}), i.e., soft
projection with bank consensus but without the layer axis. All results are
reported from a single fixed evaluation seed; the bank construction seeds
$b=1,\dots,B$ are the only stochastic component of the method, and their effect
is absorbed by the median consensus.

\subsection{Main Results}

Table~\ref{tab:main_comparison} reports the main single-category comparison
using the metrics that are consistently available in recent single-category
Dinomaly and INP-Former reports: image AUROC, pixel AP, and AUPRO. This avoids
mixing incompatible metric protocols while keeping the main comparison directly
comparable to representative memory-based, reconstruction-based, and
Transformer-based systems. ProCon achieves the strongest results on VisA and
Real-IAD among the listed methods and remains competitive on the saturated
MVTec-AD benchmark; there, where image AUROC differences among recent methods
sit within a few tenths of a point, the more informative comparison is
localization, and ProCon leads pixel AP by a clear margin ($+3.3$ over
INP-Former). Full seven-metric controlled results, including image AP/F1
and pixel AUROC/F1, are provided in the supplementary material; under that
protocol, ProCon obtains $93.2$ image AUROC and $99.0$ pixel AUROC on Real-IAD
without decoder training.

\subsection{Ablation Studies}

We isolate the contribution of each design axis under controlled settings. All
ablations use the same frozen backbone and a fixed $1\%$ coreset budget, so that
no gain can be attributed to a larger reference set; the main results use the
$5\%$ operating point motivated later in this subsection.

\begin{table*}[!t]
\centering
\footnotesize
\renewcommand{\arraystretch}{1.08}
\setlength{\tabcolsep}{3.0pt}
\begin{tabular*}{\textwidth}{@{\extracolsep{\fill}}l ccc cccc cccc}
\toprule
& \multicolumn{3}{c}{\textbf{Components}} & \multicolumn{4}{c}{\textbf{MVTec-AD}} & \multicolumn{4}{c}{\textbf{VisA}} \\
\cmidrule(lr){2-4}\cmidrule(lr){5-8}\cmidrule(lr){9-12}
\textbf{Method} & Bank & Soft & Layer & I-AUROC & P-AUROC & P-AP & AUPRO & I-AUROC & P-AUROC & P-AP & AUPRO \\
\midrule
Hard NN memory                     & --           & --           & --           & 99.5 & 97.3 & 67.4 & 92.6 & 97.8 & 96.9 & 47.3 & 92.5 \\
+ bank consensus                   & $\checkmark$ & --           & --           & 99.7 & 97.8 & 68.5 & 93.7 & 98.2 & 97.3 & 48.0 & 93.5 \\
+ soft projection (= SPM)          & $\checkmark$ & $\checkmark$ & --           & 99.7 & 98.3 & 69.6 & 94.8 & 98.5 & 98.6 & 50.3 & 95.9 \\
\textbf{+ layer consensus (= ProCon)} & $\checkmark$ & $\checkmark$ & $\checkmark$ & \textbf{99.7} & \textbf{98.6} & \textbf{73.0} & \textbf{95.7} & \textbf{99.1} & \textbf{99.0} & \textbf{52.3} & \textbf{97.0} \\
\bottomrule
\end{tabular*}
\caption{Genealogy-style ablation isolating the path from hard retrieval to
projection-consistent memory. Each row adds one design axis to the previous row;
the third row, a soft-projection memory without the layer axis, is the SPM
baseline used throughout. All rows use the same frozen backbone and a fixed
$1\%$ coreset budget. Values are rounded to one decimal.}
\label{tab:component}
\end{table*}

\paragraph{From hard retrieval to projection consensus.}
Table~\ref{tab:component} should be read as a single proof ladder in which each
row adds exactly one design axis. Starting from a plain hard nearest-neighbor
memory, adding \emph{bank consensus} shows that agreement over seed-perturbed
memories stabilizes hard retrieval (P-AP $+1.1$ on MVTec, $+0.7$ on VisA).
Adding \emph{soft projection} turns retrieval into decoder-free reconstruction
by replacing the one-anchor distance with a local normal residual (P-AUROC
$+0.5$/$+1.3$, P-AP $+1.1$/$+2.3$); the resulting single-layer method is the SPM
baseline. Finally, adding \emph{layer consensus} applies this residual objective
to independent layer-wise memories, giving the largest localization gains (P-AP
$+3.4$/$+2.0$, AUPRO $+0.9$/$+1.0$) and, on the less saturated VisA benchmark,
also lifting image-level ranking ($+0.6$); this full configuration is ProCon.
The ladder is monotone on every pixel metric on both datasets with I-AUROC never
regressing, which supports the claim that the improvement comes from the
bank/scoring redesign rather than the backbone or the memory budget.

\begin{table}[t]
\centering
\small
\renewcommand{\arraystretch}{1.15}
\setlength{\tabcolsep}{4.5pt}
\begin{tabular}{lcccc}
\toprule
\textbf{Memory} & I-AUROC & P-AUROC & P-AP & AUPRO \\
\midrule
Layer $-3$ only (deep)    & 99.4 & 98.0 & 65.4 & 93.8 \\
Layer $-6$ only           & 99.5 & 98.3 & 69.2 & 94.6 \\
Layer $-8$ only           & 99.2 & 97.9 & 69.6 & 93.9 \\
Layer $-9$ only (shallow) & 99.3 & 98.0 & \underline{72.0} & 94.1 \\
\midrule
Concat (4 layers)         & 99.6 & 98.4 & 71.2 & 95.1 \\
\textbf{ProCon} (independent) & \textbf{99.7} & \textbf{98.6} & \textbf{73.0} & \textbf{95.7} \\
\bottomrule
\end{tabular}
\caption{Single-depth memories, a concatenated four-layer memory, and the
independent four-layer consensus on MVTec-AD ($1\%$ coreset, identical banks and
readout). Layer $-3$ leads image-level detection, layer $-9$ leads pixel AP
(\underline{underlined}); the independent consensus surpasses every single depth
and the concat memory on all metrics.}
\label{tab:per_layer}
\end{table}

\paragraph{Why the depths are combined.}
The final step of the ladder---adding layer consensus---contributes the largest
localization gain yet is the least obvious to justify, so we open it up in
Table~\ref{tab:per_layer}. No depth wins on every metric: the deeper layer $-3$
gives the strongest image-level detection but the weakest pixel AP, the
shallower layer $-9$ reverses this ordering, and layers $-6$ and $-8$ sit in
between. Because the selected depths specialize in complementary metrics---and
because each is first converted into the \emph{same} soft-projection residual
before fusion---their consensus is not a compromise but an improvement,
surpassing every individual depth on all four metrics. The table also answers
the natural alternative of concatenating the same four layers into one memory
before coreset selection: under identical banks and readout, the concatenated
memory recovers only part of the multi-depth benefit (P-AP $71.2$ vs.\ $73.0$)
and trails the independent design on every metric, confirming that mixing layer
geometries before coreset selection loses information that residual-aligned
fusion preserves. This table is the concrete, per-depth counterpart of the
abstract ``+\,layer consensus'' row in Table~\ref{tab:component}. Consistent with
this, once the layer axis is in place the bank
axis adds little: a single bank per layer already reaches P-AP $72.7$ on MVTec,
and raising the bank count to $B{=}5$ adds only about $0.2$ points. The two
consensus axes are complementary in role rather than additive in magnitude,
which again points to the projection readout, not the amount of ensembled
memory, as the source of the gain.

\begin{table}[t]
\centering
\small
\renewcommand{\arraystretch}{1.1}
\setlength{\tabcolsep}{5pt}
\begin{tabular}{llccc}
\toprule
\textbf{Dataset} & \textbf{Method} & \multicolumn{3}{c}{\textbf{Pixel-AP at coreset ratio}} \\
\cmidrule(lr){3-5}
 & & $1\%$ & $5\%$ & $10\%$ \\
\midrule
\multirow{2}{*}{MVTec-AD} & SPM             & 69.6 & 70.1 & 70.1 \\
                          & \textbf{ProCon} & \textbf{73.0} & \textbf{73.5} & \textbf{73.6} \\
\midrule
\multirow{2}{*}{VisA}     & SPM             & 50.3 & 50.3 & 50.3 \\
                          & \textbf{ProCon} & \textbf{52.3} & \textbf{52.3} & \textbf{52.3} \\
\bottomrule
\end{tabular}
\caption{Coreset-budget sweep (pixel AP). ProCon at $1\%$ already exceeds SPM at
$10\%$ on both datasets, and both methods saturate at $5\%$. Full eight-metric
budget tables are in the supplementary material.}
\label{tab:budget}
\end{table}

\paragraph{Memory design versus memory budget.}
Table~\ref{tab:budget} separates the readout from the memory size. ProCon
dominates the single-layer SPM baseline at every coreset ratio, and ProCon with a
$1\%$ coreset already outperforms SPM with a $10\%$ coreset on both benchmarks,
so the gain cannot be explained by storing more anchors. Both methods saturate at
$5\%$, which motivates the $5\%$ operating point used on MVTec-AD and VisA in the
main comparison. Layer-pool selection, fusion-rule selection, the full
bank-count sweep, seven-metric ablations, and negative design results are
provided in the supplementary material.

\subsection{Real-IAD Generalization}

Real-IAD is substantially larger and more diverse than MVTec-AD and VisA, and
image-level scores are less saturated. ProCon is applied to all $30$ categories
with the same layer pool and $1\%$ coreset, and without retuning the projection
rule. The resulting $93.2$ image AUROC and $97.2$ AUPRO indicate that the
projection-consistency readout transfers beyond small benchmarks. Pixel AP is
lower than on MVTec-AD and VisA because many Real-IAD defects occupy very small
regions, so precision-recall metrics are sensitive to tiny false-positive areas;
the high pixel AUROC ($99.0$) and AUPRO ($97.2$), however, show that ProCon still
ranks anomalous pixels reliably over large images. The gains are not uniform
across categories: rigid, well-structured parts such as rolled strip bases,
switches, and zippers approach the saturation seen on MVTec-AD (image AUROC
$99.0$--$99.5$), while fine-texture categories with very small defects, such as
toy bricks and mints, remain the hardest ($82$--$86$), suggesting that the
remaining headroom lies in texture-dominated normality rather than in the
readout. Per-category results for all $30$ categories are provided in the
supplementary material.

\subsection{Qualitative Analysis}

\begin{figure}[t]
\centering
\includegraphics[width=\linewidth]{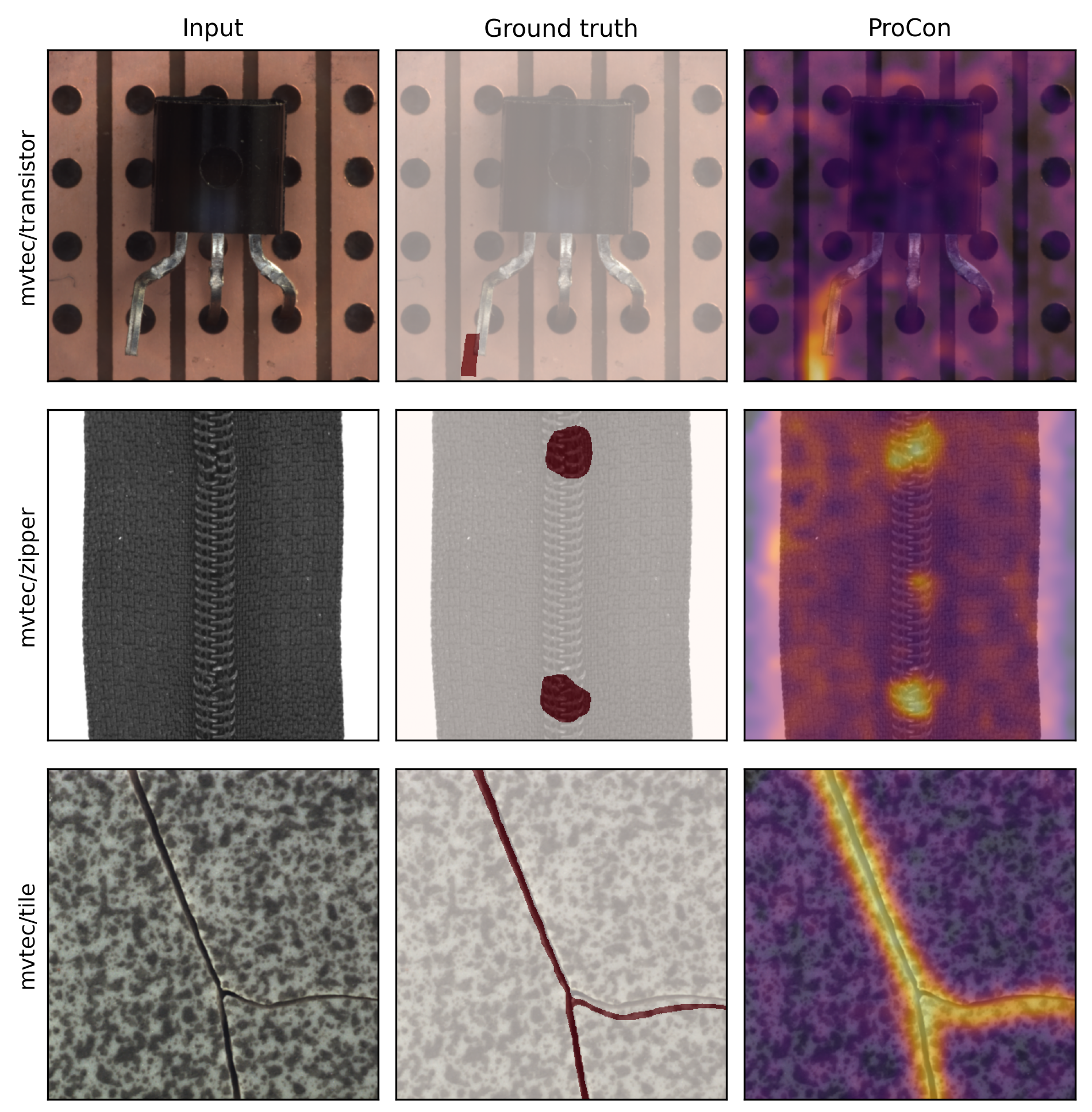}
\caption{Compact qualitative examples. ProCon localizes structural and texture
defects; extended qualitative maps and layer-wise residuals are in the
supplement.}
\label{fig:qualitative}
\end{figure}

Figure~\ref{fig:qualitative} shows representative anomaly maps for structural and
texture defects. The examples match the quantitative trend: projection residuals
highlight small defects without training a decoder.

\section{Discussion}

\paragraph{Why projection residuals help.}
The main empirical pattern is that ProCon improves localization more
consistently than image-level detection. This is expected: image-level AUROC is
already near saturation on MVTec-AD, whereas pixel-level AP and AUPRO remain
sensitive to faint boundaries, thin cracks, and small missing parts.
Nearest-neighbor scoring can suppress exactly this evidence, because a defective
patch that happens to lie close to one normal anchor receives a small distance
and is accepted as normal. Soft projection instead asks whether several nearby
anchors \emph{agree} on a normal explanation: if the local neighborhood is
geometrically inconsistent with the test patch, the weighted normal estimate
cannot reconstruct it and the residual stays high. The pattern is clearest where
image-level scores are least saturated: on VisA the readout change also lifts
image-level ranking, and on Real-IAD ProCon improves image AUROC by more than a
point over the strongest reconstruction-based baselines while keeping pixel
ranking metrics high. The readout therefore behaves like reconstruction while
retaining the simplicity of a fixed memory bank, and because the projected
estimate is confined to the neighborhood spanned by stored normal anchors, the
residual is tied directly to observed normal variation rather than to a learned
decoder that might reconstruct anomalies too well.

\paragraph{Why consensus is applied after alignment, not before.}
A central design choice is the order of operations. A tempting alternative is to
concatenate multi-layer features and build a single memory bank, but this mixes
feature spaces whose distance scales and normal geometries differ, and a coreset
selected in the concatenated space can over-represent high-variance layers while
under-representing layers useful for fine localization;
Table~\ref{tab:per_layer} quantifies this cost directly. ProCon instead enforces
a three-step order: construct memories independently, convert each memory
response into a projection residual, and only then aggregate. The residuals then
carry the same semantic meaning across banks and layers---unexplained distance
from a local normal projection. Bank consensus removes memory-sampling noise and
layer consensus combines already-aligned evidence, so the improvement in
Table~\ref{tab:component} is not a generic ensemble effect but a consequence of
aligning each memory response into a comparable residual before fusion.

\paragraph{Why memory size alone is insufficient.}
A denser coreset gives more normal anchors, but density does not solve the
false-normal problem; if anything, a larger memory makes it easier for a
defective patch to find at least one nearby anchor. ProCon changes the decision
criterion rather than the memory capacity: a patch must be supported by a local
neighborhood, and that support must remain stable across seed-perturbed memories
and across depths. The budget sweep in Table~\ref{tab:budget} makes this
concrete, since ProCon at a $1\%$ coreset already outperforms the single-layer
SPM baseline at a $10\%$ coreset on both datasets: the readout geometry, not the
number of stored anchors, is responsible for the improvement. This is also why
the method is easy to audit: because every design choice is fixed across
datasets, a performance change can be traced to memory construction, projection
residuals, or consensus, rather than to dataset-specific calibration.

\section{Conclusion}

We introduced ProCon, a training-free projection-consistency memory framework
that replaces hard nearest-neighbor scoring with soft local normal projection
and uses the projection residual as decoder-free anomaly evidence, stabilized by
bank-wise and layer-wise consensus. Across MVTec-AD, VisA, and Real-IAD, ProCon
delivers consistent image- and pixel-level gains, and controlled ablations
attribute the improvement to the projection-consistency readout rather than to
larger memories. The method occupies a middle ground between memory retrieval
and learned reconstruction, obtaining reconstruction-style residuals while
retaining the simplicity of fixed normal memory. The broader design rule is
simple: memory should be treated not
as a set of stored exemplars but as a normal support set queried for local
explainability---a readout question that remains the same as encoders improve.

Several limitations point to natural extensions. ProCon trades decoder training
for repeated memory retrieval, evaluating $20$ layer-bank memories per image;
high-throughput inspection would benefit from approximate nearest-neighbor
search or compressed coresets, which do not change the scoring objective. The
layer pool is fixed across categories for reproducibility and may not be optimal
for every object type, so adaptive layer selection is a natural but
overfitting-prone direction. Finally, like other normal-memory systems ProCon
assumes that the nominal set defines reliable normal support; repeated abnormal
patterns in training images could be absorbed as normal anchors, which memory
cleaning or confidence-weighted coreset construction could address.

\clearpage
\bibliography{aaai2027}

\clearpage
\begin{center}
{\LARGE\bf Supplementary Material}\\[0.6em]
{\large ProCon: Projection-Consistency Memory for Training-Free Anomaly Detection}
\end{center}
\vspace{0.5em}

\section{Exact Implementation Details}

This section provides reference-grade implementation details for ProCon. Unless otherwise specified, ProCon uses a frozen DINOv2 ViT-B/14 backbone, the layer pool $\{-3,-6,-8,-9\}$, $B=5$ seed-perturbed memory banks per layer, soft projection with $k=5$, bank-wise median consensus, layer-wise mean consensus, no per-image normalization, and top-mean image-level readout with ratio $0.005$.

\subsection{Feature Extraction}

All images are resized to $392 \times 392$. With DINOv2 ViT-B/14, this produces a $28 \times 28$ patch grid, corresponding to $P=784$ patch tokens. We extract patch tokens from selected transformer layers $\mathcal{L}=\{-3,-6,-8,-9\}$, where negative indices count backward from the final transformer block. The CLS token is removed and only patch tokens are used for memory construction and anomaly scoring. Patch tokens are L2-normalized per layer before the fixed projection described below.

All selected layers share the same spatial grid, so layer-wise residual maps are spatially aligned without interpolation. This avoids resolution confounds when fusing residual maps across representation depths. All selected layers are extracted from a single frozen forward pass through the backbone.

\subsection{Fixed Random Projection}

For each selected layer $\ell$, we apply a fixed layer-specific random projection before coreset construction and test-time scoring. The projection maps DINOv2 patch features to a lower-dimensional memory space and is never trained. The same projected feature space is used for both training-memory construction and test-time soft projection, so train and test patches are compared in the same geometry.

\begin{table}[t]
\centering
\begin{tabular}{ll}
\toprule
Component & Value \\
\midrule
Raw feature dimension & 768 \\
Projected dimension (scoring space) & 512 \\
Distance sub-projection (coreset only) & 192 \\
Projection type & fixed linear, no bias \\
Projection seed & 42 \\
Trainable parameters & none \\
\bottomrule
\end{tabular}
\caption{Fixed projection configuration. The 192-dimensional sub-projection is used only to accelerate distance computation during coreset selection; stored anchors and all scoring operate in the 512-dimensional space.}
\label{tab:s_projection}
\end{table}

\subsection{Layer-wise Coreset Construction}

For each layer $\ell$ and bank index $b$, ProCon constructs an independent coreset memory $M_{\ell,b}$ from normal training patches by greedy farthest-point (k-center) selection at the target coreset ratio. The coreset is selected independently per layer rather than from concatenated multi-layer features, which preserves depth-specific normal geometry before residual fusion.

We use $B=5$ seed-perturbed memory banks for each selected layer. Different banks use different random seeds during coreset construction, producing slightly different normal reference samples; these banks feed the bank-wise median consensus at test time.

For efficiency, the pairwise distances that drive k-center selection are computed in a fixed 192-dimensional sub-projection of the 512-dimensional scoring space, while the stored memory anchors remain the full 512-dimensional vectors. This accelerates selection without changing the feature geometry used for test-time soft projection.

\begin{itemize}
\item Coreset selection is performed separately for each layer and each bank.
\item The bank index controls the random seed, producing seed-perturbed memories.
\item Distances for coreset selection are computed in the 192-dimensional sub-projection; stored anchors are 512-dimensional.
\item No parameters are learned during memory construction.
\end{itemize}

\subsection{Soft Projection and Auto Temperature}

For each test patch feature $z_{\ell,p}$, ProCon retrieves its $k$ nearest normal memory vectors from $M_{\ell,b}$ and writes $\mathcal{N}_k=\mathrm{kNN}(z_{\ell,p},M_{\ell,b})$. The soft projection and residual are
\[
\hat z_{\ell,b,p}
=
\sum_{m_j\in\mathcal{N}_k}
w_j m_j,
\qquad
r_{\ell,b,p}
=
\|z_{\ell,p}-\hat z_{\ell,b,p}\|_2,
\]
with distance-based weights
\[
w_j=
\frac{
\exp\left(-\|z_{\ell,p}-m_j\|_2^2/\tau_{\ell,b}\right)
}{
\sum_{m_t\in\mathcal{N}_k}
\exp\left(-\|z_{\ell,p}-m_t\|_2^2/\tau_{\ell,b}\right)
}.
\]

The temperature $\tau_{\ell,b}$ is set per test image and per memory bank from local distance statistics:
\[
\tau_{\ell,b}
=
\mathrm{median}
\left(
\left\{
\|z_{\ell,p}-m_j\|_2^2:
m_j\in \mathcal{N}_k
\right\}_{p,j}
\right),
\]
where, for efficiency, $p$ ranges over a fixed-size random subsample of $512$ patches of the current test image rather than all $P$ patches. This avoids using a fixed temperature across datasets, categories, layers, and memory banks: a fixed temperature can either flatten the softmax too much or make it collapse to nearly nearest-neighbor behavior, depending on the feature-distance scale.

\subsection{Two-axis Consensus and Readout}

For each layer $\ell$, ProCon first aggregates residual maps over seed-perturbed memory banks by a median:
\[
S_{\ell}(p)
=
\operatorname*{median}_{b=1}^{B} r_{\ell,b,p}.
\]
The final anomaly map is then obtained by a mean over selected layers:
\[
S_{\mathrm{map}}(p)
=
\frac{1}{|\mathcal{L}|}
\sum_{\ell\in\mathcal{L}} S_\ell(p).
\]
We do not apply per-image normalization before layer fusion. Pixel-level metrics are computed directly from $S_{\mathrm{map}}$.

For image-level detection, the image score is computed by top-mean pooling:
\[
S_{\mathrm{img}}
=
\frac{1}{|\mathcal{T}_{\rho}|}
\sum_{p\in\mathcal{T}_{\rho}}
S_{\mathrm{map}}(p),
\]
where $\mathcal{T}_{\rho}$ contains the top $\rho=0.005$ fraction of patch scores.

\subsection{Full Procedure}

Algorithm~\ref{alg:s_procon} summarizes the complete ProCon pipeline referenced from the main paper.

\begin{algorithm}[t]
\caption{ProCon: Projection-Consistency Memory}
\label{alg:s_procon}
\begin{algorithmic}[1]
\REQUIRE Normal images $\mathcal{D}_{\mathrm{train}}$, frozen backbone $f_\theta$, layers $\mathcal{L}$, banks $B$, coreset ratio $\rho_m$
\ENSURE Anomaly map $S_{\mathrm{map}}$, image score $S_{\mathrm{img}}$
\STATE \textbf{Memory construction (per category)}
\FOR{each layer $\ell\in\mathcal{L}$}
\STATE Extract normal patch features $F_\ell(\mathcal{D}_{\mathrm{train}})$
\STATE Apply fixed projection $G_\ell$ (L2 normalize, project to 512-d)
\FOR{each bank $b=1,\dots,B$}
\STATE $M_{\ell,b}\leftarrow$ k-center coreset with seed $b$, ratio $\rho_m$
\ENDFOR
\ENDFOR
\STATE \textbf{Inference (per test image $x$)}
\STATE Extract $F_\ell(x)$ for all $\ell\in\mathcal{L}$ in one forward pass
\FOR{each layer $\ell\in\mathcal{L}$}
\FOR{each bank $b=1,\dots,B$}
\STATE Compute $\tau_{\ell,b}$ from a 512-patch subsample of $x$
\STATE Retrieve $\mathcal{N}_k$ from $M_{\ell,b}$; soft-project each patch
\STATE $r_{\ell,b,p}\leftarrow\|z_{\ell,p}-\hat z_{\ell,b,p}\|_2$
\ENDFOR
\STATE $S_\ell(p)\leftarrow\operatorname{median}_b\, r_{\ell,b,p}$
\ENDFOR
\STATE $S_{\mathrm{map}}(p)\leftarrow|\mathcal{L}|^{-1}\sum_{\ell}S_\ell(p)$
\STATE $S_{\mathrm{img}}\leftarrow\mathrm{TopMean}_{0.005}(S_{\mathrm{map}})$
\RETURN $S_{\mathrm{map}},S_{\mathrm{img}}$
\end{algorithmic}
\end{algorithm}

\subsection{Complexity}

For each test image, ProCon performs $|\mathcal{L}|B$ soft-projection evaluations. With $|\mathcal{L}|=4$ and $B=5$, this gives $20$ layer-bank evaluations per image. If each memory bank stores $|M|$ vectors, each image has $P$ patch tokens, and the projected feature dimension is $D$, the dominant retrieval cost is
\[
O(|\mathcal{L}|\,B\,P\,|M|\,D).
\]
This retrieval cost replaces decoder training and learned reconstruction with fixed memory operations. The dominant offline cost is the greedy coreset selection, not scoring.

\section{Full Experimental Protocol}

The final ProCon recipe was obtained through a controlled multi-phase protocol. Each phase fixes the decisions made in previous phases and varies only one factor at a time. This prevents layer choice, fusion rule, normalization, and readout from being confounded.

\subsection{Full Seven-Metric Controlled Results}

Table~\ref{tab:s_seven_metric} reports the seven-metric controlled results referenced in the main paper: ProCon against the SPM baseline under identical backbone, budget, and readout, so that every difference between the rows is attributable to the layer-consensus axis. On Real-IAD only the final ProCon configuration was evaluated.

\begin{table*}[t]
\centering
\setlength{\tabcolsep}{4pt}
\renewcommand{\arraystretch}{1.05}
\begin{tabular}{llccccccc}
\toprule
& & \multicolumn{3}{c}{Image-level} & \multicolumn{4}{c}{Pixel-level} \\
\cmidrule(lr){3-5}\cmidrule(lr){6-9}
Dataset & Method & AUROC & AP & F1 & AUROC & AP & F1 & AUPRO \\
\midrule
MVTec-AD & SPM & 99.73 & 99.92 & \textbf{99.45} & 98.46 & 70.08 & 68.91 & 95.11 \\
MVTec-AD & \textbf{ProCon} & \textbf{99.75} & \textbf{99.92} & 99.32 & \textbf{98.69} & \textbf{73.47} & \textbf{70.92} & \textbf{95.86} \\
\midrule
VisA & SPM & 98.57 & 98.79 & 96.12 & 98.67 & 50.34 & 54.02 & 96.13 \\
VisA & \textbf{ProCon} & \textbf{99.19} & \textbf{99.30} & \textbf{97.46} & \textbf{99.07} & \textbf{52.28} & \textbf{54.72} & \textbf{97.03} \\
\midrule
Real-IAD & \textbf{ProCon} & 93.15 & 91.15 & 84.77 & 99.04 & 49.35 & 51.49 & 97.19 \\
\bottomrule
\end{tabular}
\caption{Full seven-metric controlled results at the final operating points ($5\%$ coreset for MVTec-AD and VisA, $1\%$ for Real-IAD). SPM denotes the single-layer soft-projection memory baseline, i.e., ProCon without the layer-consensus axis.}
\label{tab:s_seven_metric}
\end{table*}

\subsection{Four-phase Protocol}

\begin{table*}[t]
\centering
\scriptsize
\renewcommand{\arraystretch}{1.05}
\resizebox{\textwidth}{!}{%
\begin{tabular}{lllll}
\toprule
Phase & Question & Varies & Fixed & Validation Set \\
\midrule
1 & Which layers are useful? & single layer $\ell$ & soft projection scoring & MVTec-AD 15 categories \\
2 & How should layers be fused? & aggregation, normalization, readout & temporary pool $\{-3,-4,-6,-8,-9\}$ & bottom-3 subset \\
3 & Which layer pool survives? & layer pool & mean, no normalization, top-mean 0.005 & bottom-3 subset \\
4 & Does the pool generalize? & promoted pools & fixed recipe & full MVTec-AD and VisA \\
\bottomrule
\end{tabular}}
\caption{Controlled protocol used to select the final ProCon recipe. Phase 2 uses a temporary five-layer pool only to select the fusion rule; the final four-layer pool is selected in Phase 3 and promoted in Phase 4.}
\label{tab:s_protocol}
\end{table*}

\subsection{Validation Subset and Sanity Gate}

For rapid controlled sweeps, we use a bottom-3 validation subset consisting of \emph{transistor}, \emph{pill}, and \emph{zipper}. These categories expose instability in image-level detection and localization. We also use a sanity gate based on transistor image AUROC: recipes with transistor I-AUROC below $0.99$ are considered unstable and are not promoted, even if they achieve high pixel AP on the validation subset.

\section{Additional Component Analysis}

This section expands the component analysis from the main paper. All variants in this section use the same frozen backbone and the same $1\%$ coreset budget. The purpose is to isolate the effect of each design axis, not to report the final benchmark setting.

\subsection{Controlled Genealogy}

\begin{table*}[t]
\centering
\setlength{\tabcolsep}{4pt}
\begin{tabular}{lcccccccc}
\toprule
& \multicolumn{4}{c}{MVTec-AD} & \multicolumn{4}{c}{VisA} \\
\cmidrule(lr){2-5}\cmidrule(lr){6-9}
Variant & I-AUROC & P-AUROC & P-AP & AUPRO & I-AUROC & P-AUROC & P-AP & AUPRO \\
\midrule
Hard NN memory & 99.46 & 97.29 & 67.35 & 92.57 & 97.75 & 96.88 & 47.28 & 92.45 \\
+ Bank consensus & 99.71 & 97.82 & 68.45 & 93.72 & 98.22 & 97.33 & 48.01 & 93.45 \\
+ Soft projection (SPM) & 99.71 & 98.33 & 69.55 & 94.81 & 98.50 & 98.59 & 50.28 & 95.91 \\
\textbf{+ Layer consensus (ProCon)} & \textbf{99.71} & \textbf{98.62} & \textbf{72.98} & \textbf{95.66} & \textbf{99.10} & \textbf{99.03} & \textbf{52.29} & \textbf{96.95} \\
\bottomrule
\end{tabular}
\caption{Controlled genealogy from hard retrieval to projection-consistent memory under a fixed $1\%$ coreset budget, reported at two decimals. Row names match the ablation ladder in the main paper.}
\label{tab:s_genealogy}
\end{table*}

The progression shows that the gain is not caused by a single implementation detail. Bank consensus improves hard retrieval by reducing memory-sampling sensitivity. Soft projection turns nearest-neighbor lookup into decoder-free reconstruction. ProCon further improves localization by applying the same residual objective to independent layer-wise memories.

\subsection{Layer Independence: Concatenated versus Independent Memory}

The main paper reports that independent per-layer memories outperform a concatenated single memory under the final operating point. Table~\ref{tab:s_concat} gives the full eight-metric comparison. Both rows use full MVTec-AD, the final four-layer pool, a $1\%$ coreset, $B=5$ seed banks, soft-projection median, and top-mean $0.005$ readout; the only difference is whether the four layers are concatenated into one memory before coreset selection or kept as independent per-layer memories fused after residual alignment.

\begin{table*}[t]
\centering
\setlength{\tabcolsep}{4pt}
\begin{tabular}{lcccccccc}
\toprule
Memory & I-AUROC & I-AP & I-F1 & P-AUROC & P-AP & P-F1 & AUPRO & PRO \\
\midrule
Concat (single memory) & 99.56 & 99.83 & 99.06 & 98.39 & 71.22 & 69.67 & 95.06 & 90.77 \\
\textbf{ProCon (independent)} & \textbf{99.71} & \textbf{99.90} & \textbf{99.24} & \textbf{98.62} & \textbf{72.98} & \textbf{70.56} & \textbf{95.66} & \textbf{92.74} \\
\bottomrule
\end{tabular}
\caption{Concatenated single memory versus independent per-layer memories at the final operating point (full MVTec-AD, four-layer pool, $1\%$ coreset, identical banks and readout). Independent memory wins on all eight metrics, with the largest gaps in localization (P-AP $+1.76$, AUPRO $+0.60$, PRO $+1.97$). Note that this concat result is not comparable to the concat entry in the Phase-2 fusion sweep (Table~\ref{tab:s_fusion}), which uses the bottom-3 subset and a temporary five-layer pool.}
\label{tab:s_concat}
\end{table*}

\subsection{Bank-count Sweep}

How much does the bank-consensus axis contribute once the layer-consensus axis is already present? Table~\ref{tab:s_banks} sweeps the number of seed-perturbed banks $B$ from $1$ to $5$ on full MVTec-AD, holding everything else at the final configuration. $B=1$ means a single coreset per layer with no bank consensus.

\begin{table*}[t]
\centering
\setlength{\tabcolsep}{4pt}
\begin{tabular}{ccccccccc}
\toprule
$B$ & I-AUROC & I-AP & I-F1 & P-AUROC & P-AP & P-F1 & AUPRO & PRO \\
\midrule
1 & 99.68 & 99.88 & \textbf{99.28} & 98.57 & 72.74 & 70.40 & 95.53 & 92.16 \\
2 & 99.70 & 99.88 & 99.21 & 98.58 & 72.92 & 70.55 & 95.58 & 90.93 \\
3 & 99.70 & 99.89 & 99.21 & 98.60 & 72.89 & 70.53 & 95.61 & 90.44 \\
4 & 99.70 & 99.89 & 99.19 & 98.61 & \textbf{73.00} & \textbf{70.57} & 95.65 & 91.90 \\
\textbf{5} & \textbf{99.71} & \textbf{99.90} & 99.24 & \textbf{98.62} & 72.98 & 70.56 & \textbf{95.66} & \textbf{92.74} \\
\bottomrule
\end{tabular}
\caption{Bank-count sweep on full MVTec-AD at the final configuration ($1\%$ coreset, four-layer pool). Ranking metrics increase monotonically with $B$ but the gains are small and saturating; a single bank per layer already reaches P-AP $72.74$, well above the SPM reference ($69.55$, Table~\ref{tab:s_genealogy}).}
\label{tab:s_banks}
\end{table*}

The ranking metrics (I-AUROC, P-AUROC, AUPRO) increase monotonically with $B$, confirming that bank consensus still stabilizes the residual, but the total gain from $B=1$ to $B=5$ is only about $0.05$--$0.13$ points and P-AP is essentially flat. Crucially, $B=1$ already exceeds the SPM reference by a wide margin. The interpretation is that the two consensus axes are partially redundant stabilizers: averaging four independent layer maps is itself a strong variance-reduction ensemble, so once the layer axis is in place there is little residual noise left for additional banks to suppress. ProCon therefore keeps $B=5$ as a cheap safety default rather than a load-bearing component; the localization gain comes from the layer axis.

\section{Layer-wise Analysis and Pool Selection}

This section provides the layer-wise diagnostic results and the pool-selection procedure. Importantly, single-layer profiling is diagnostic rather than decisive: the final pool is selected by contribution to multi-layer residual consensus, not by isolated single-layer performance.

\subsection{Single-layer Profiling}

\begin{table}[t]
\centering
\begin{tabular}{lcccc}
\toprule
Layer & I-AUROC & P-AUROC & P-AP & AUPRO \\
\midrule
$-1$ & 99.24 & 96.95 & 57.44 & 90.94 \\
$-2$ & 99.28 & 97.32 & 57.63 & 92.02 \\
$-3$ & 99.41 & 97.98 & 65.43 & 93.77 \\
$-4$ & \textbf{99.57} & 98.12 & 66.12 & 94.14 \\
$-6$ & 99.45 & \textbf{98.26} & 69.19 & \textbf{94.59} \\
$-8$ & 99.15 & 97.92 & 69.61 & 93.91 \\
$-9$ & 99.26 & 97.96 & \textbf{72.02} & 94.13 \\
$-10$ & 97.62 & 96.60 & 67.60 & 90.85 \\
$-12$ & 80.19 & 81.36 & 38.61 & 62.78 \\
\bottomrule
\end{tabular}
\caption{Single-layer profiling on MVTec-AD. Layer $-9$ gives the highest P-AP, layer $-6$ gives the highest P-AUROC and AUPRO, and layer $-4$ gives the highest single-layer I-AUROC.}
\label{tab:s_single_layer}
\end{table}

The table shows that different depths specialize in different metrics. Layer $-9$ is strongest in P-AP, layer $-6$ is strongest in P-AUROC and AUPRO, and layer $-4$ is strongest in single-layer I-AUROC. However, the final pool is not obtained by selecting the individually best layer for each metric. Instead, we select layers by their contribution inside the multi-layer residual consensus. Although $-4$ is strong as a single layer, adding it to the temporary five-layer pool lowers P-AP compared with the final four-layer pool, so it is excluded.

\subsection{Fusion Rule Selection on a Temporary Pool}

We choose the fusion rule using a temporary five-layer pool $\{-3,-4,-6,-8,-9\}$. This experiment is not the final ProCon configuration. Its only purpose is to select the aggregation and normalization strategy before layer-pool pruning.

\begin{table}[t]
\centering
\begin{tabular}{lccc}
\toprule
Variant & I-AUROC & P-AP & AUPRO \\
\midrule
Concat single-bank & 99.75 & 68.54 & 90.34 \\
Layer-wise memory, mean & \textbf{99.79} & \textbf{70.44} & \textbf{91.69} \\
Layer-wise memory, median & 99.82 & 69.49 & 91.18 \\
Layer-wise memory, max & 99.57 & 66.24 & 89.43 \\
Robust per-image norm & 91.7 & 41.0 & 67.0 \\
\bottomrule
\end{tabular}
\caption{Fusion rule selection on the bottom-3 subset with the temporary five-layer pool $\{-3,-4,-6,-8,-9\}$. Mean aggregation without per-image normalization is selected. The concat entry here is under this Phase-2 protocol and is not comparable to the final-operating-point concat comparison in Table~\ref{tab:s_concat}.}
\label{tab:s_fusion}
\end{table}

\subsection{Layer-pool Pruning}

\begin{table}[t]
\centering
\begin{tabular}{lccc}
\toprule
Layer Pool & I-AUROC & P-AP & AUPRO \\
\midrule
$\{-8,-9\}$ & 99.28 & \textbf{74.56} & 89.89 \\
$\{-6,-8,-9\}$ & 99.58 & 73.70 & 91.07 \\
$\{-6,-9\}$ & 99.54 & 73.01 & 91.18 \\
$\{-3,-6,-8,-9\}$ & 99.75 & 72.21 & \textbf{91.72} \\
$\{-3,-4,-6,-8,-9\}$ & \textbf{99.79} & 70.44 & 91.69 \\
\bottomrule
\end{tabular}
\caption{Layer-pool pruning on the bottom-3 validation subset. The final pool $\{-3,-6,-8,-9\}$ provides the best robustness-localization trade-off and is promoted to the full datasets.}
\label{tab:s_layer_pool}
\end{table}

The two-layer pool $\{-8,-9\}$ achieves the highest P-AP on the validation subset, but it loses AUPRO and image-level robustness. The five-layer pool $\{-3,-4,-6,-8,-9\}$ gives the highest validation I-AUROC but lower P-AP than the final four-layer pool. Therefore, the final pool $\{-3,-6,-8,-9\}$ is selected by promotion behavior rather than by any isolated validation metric.

\subsection{Full-dataset Promotion}

The top candidate pools from the validation subset are promoted to full MVTec-AD and VisA. The final pool must improve P-AUROC, P-AP, and AUPRO over SPM while preserving image-level AUROC.

\begin{table}[t]
\centering
\begin{tabular}{lcccc}
\toprule
Pool & I-AUROC & P-AUROC & P-AP & AUPRO \\
\midrule
$\{-3,-6,-8,-9\}$ & \textbf{99.71} & \textbf{98.62} & 72.98 & \textbf{95.66} \\
$\{-6,-9\}$ & 99.58 & 98.52 & \textbf{73.29} & 95.43 \\
$\{-6,-8,-9\}$ & 99.55 & 98.46 & 72.87 & 95.29 \\
$\{-8,-9\}$ & 99.37 & 98.18 & 72.28 & 94.63 \\
SPM & 99.71 & 98.33 & 69.55 & 94.81 \\
\bottomrule
\end{tabular}
\caption{Full MVTec-AD promotion results at the controlled $1\%$ coreset setting.}
\label{tab:s_promotion_mvtec}
\end{table}

\begin{table}[t]
\centering
\begin{tabular}{lcccc}
\toprule
Pool & I-AUROC & P-AUROC & P-AP & AUPRO \\
\midrule
$\{-3,-6,-8,-9\}$ & \textbf{99.10} & \textbf{99.03} & \textbf{52.29} & \textbf{96.95} \\
$\{-6,-9\}$ & 98.99 & 98.49 & 49.88 & 95.70 \\
$\{-6,-8,-9\}$ & 98.94 & 98.33 & 49.50 & 95.39 \\
$\{-8,-9\}$ & 98.60 & 97.51 & 48.01 & 93.76 \\
SPM & 98.50 & 98.59 & 50.28 & 95.91 \\
\bottomrule
\end{tabular}
\caption{Full VisA promotion results at the controlled $1\%$ coreset setting. The final pool $\{-3,-6,-8,-9\}$ is the only promoted pool that improves all pixel metrics over SPM.}
\label{tab:s_promotion_visa}
\end{table}

\section{Full Coreset Budget Study}

We evaluate ProCon and SPM at $1\%$, $5\%$, and $10\%$ coreset ratios on MVTec-AD and VisA. The purpose is to test whether ProCon's gain comes from memory design rather than memory size.

\begin{table*}[t]
\centering
\begin{tabular}{llcccccccc}
\toprule
Method & Budget & I-AUROC & I-AP & I-F1 & P-AUROC & P-AP & P-F1 & AUPRO & PRO \\
\midrule
ProCon & $1\%$ & 99.71 & 99.90 & 99.24 & 98.62 & 72.98 & 70.56 & 95.66 & 92.74 \\
ProCon & $5\%$ & 99.75 & 99.92 & 99.32 & 98.69 & 73.47 & 70.92 & 95.86 & \textbf{93.38} \\
ProCon & $10\%$ & \textbf{99.76} & \textbf{99.93} & 99.40 & \textbf{98.70} & \textbf{73.55} & \textbf{70.99} & \textbf{95.88} & 92.73 \\
SPM & $1\%$ & 99.71 & 99.90 & 99.38 & 98.33 & 69.55 & 68.43 & 94.81 & 90.56 \\
SPM & $5\%$ & 99.73 & 99.92 & 99.45 & 98.46 & 70.08 & 68.91 & 95.11 & 92.19 \\
SPM & $10\%$ & 99.72 & 99.91 & \textbf{99.48} & 98.48 & 70.10 & 68.92 & 95.15 & 92.06 \\
\bottomrule
\end{tabular}
\caption{Coreset budget sweep on MVTec-AD. ProCon outperforms SPM across all budgets; ProCon at $1\%$ already exceeds SPM at $10\%$ on P-AP.}
\label{tab:s_budget_mvtec}
\end{table*}

\begin{table*}[t]
\centering
\begin{tabular}{llcccccccc}
\toprule
Method & Budget & I-AUROC & I-AP & I-F1 & P-AUROC & P-AP & P-F1 & AUPRO & PRO \\
\midrule
ProCon & $1\%$ & 99.10 & 99.24 & 97.13 & 99.03 & 52.29 & \textbf{54.93} & 96.95 & \textbf{90.30} \\
ProCon & $5\%$ & \textbf{99.19} & \textbf{99.30} & \textbf{97.46} & 99.07 & 52.28 & 54.72 & 97.03 & 89.59 \\
ProCon & $10\%$ & 99.15 & 99.27 & 97.42 & \textbf{99.08} & \textbf{52.32} & 54.68 & \textbf{97.04} & 89.50 \\
SPM & $1\%$ & 98.50 & 98.77 & 96.04 & 98.59 & 50.28 & 53.95 & 95.91 & 88.12 \\
SPM & $5\%$ & 98.57 & 98.79 & 96.12 & 98.67 & 50.34 & 54.02 & 96.13 & 87.15 \\
SPM & $10\%$ & 98.49 & 98.71 & 96.09 & 98.68 & 50.33 & 53.90 & 96.13 & 87.33 \\
\bottomrule
\end{tabular}
\caption{Coreset budget sweep on VisA. ProCon remains stronger than SPM across all budgets.}
\label{tab:s_budget_visa}
\end{table*}

\section{Full Per-category Results}

This section reports full per-category results for ProCon. We report all eight metrics: image AUROC, image AP, image F1, pixel AUROC, pixel AP, pixel F1, AUPRO, and PRO.

\subsection{MVTec-AD Final Results at $5\%$ Coreset}

\begin{table*}[t]
\centering
\begin{tabular}{lcccccccc}
\toprule
Category & I-AUROC & I-AP & I-F1 & P-AUROC & P-AP & P-F1 & AUPRO & PRO \\
\midrule
bottle & 100.00 & 100.00 & 100.00 & 99.15 & 89.35 & 82.64 & 97.18 & 96.03 \\
cable & 99.81 & 99.89 & 98.36 & 98.65 & 77.69 & 72.86 & 95.66 & 94.35 \\
capsule & 98.76 & 99.71 & 98.64 & 98.88 & 61.88 & 58.05 & 96.31 & 90.81 \\
carpet & 100.00 & 100.00 & 100.00 & 99.55 & 79.28 & 76.03 & 98.50 & 93.54 \\
grid & 100.00 & 100.00 & 100.00 & 99.63 & 66.36 & 63.41 & 98.77 & 97.46 \\
hazelnut & 100.00 & 100.00 & 100.00 & 99.54 & 82.45 & 78.96 & 98.47 & 92.33 \\
leather & 100.00 & 100.00 & 100.00 & 99.58 & 61.69 & 60.25 & 98.60 & 95.33 \\
metal\_nut & 100.00 & 100.00 & 100.00 & 97.93 & 82.99 & 87.17 & 93.16 & 93.28 \\
pill & 99.40 & 99.90 & 98.94 & 97.25 & 74.71 & 69.11 & 92.26 & 96.53 \\
screw & 98.77 & 99.58 & 97.12 & 99.47 & 64.71 & 60.92 & 98.41 & 95.44 \\
tile & 100.00 & 100.00 & 100.00 & 98.32 & 75.81 & 79.08 & 94.41 & 87.21 \\
toothbrush & 100.00 & 100.00 & 100.00 & 99.32 & 62.89 & 68.66 & 97.72 & 92.61 \\
transistor & 99.88 & 99.82 & 97.56 & 96.54 & 71.88 & 66.84 & 89.19 & 86.49 \\
wood & 99.65 & 99.89 & 99.17 & 97.98 & 78.42 & 71.53 & 93.91 & 93.95 \\
zipper & 100.00 & 100.00 & 100.00 & 98.58 & 71.91 & 68.25 & 95.42 & 95.27 \\
\textbf{Mean} & \textbf{99.75} & \textbf{99.92} & \textbf{99.32} & \textbf{98.69} & \textbf{73.47} & \textbf{70.92} & \textbf{95.86} & \textbf{93.38} \\
\bottomrule
\end{tabular}
\caption{MVTec-AD per-category results for ProCon at the final $5\%$ coreset setting.}
\label{tab:s_mvtec_percat}
\end{table*}

\subsection{VisA Final Results at $5\%$ Coreset}

\begin{table*}[t]
\centering
\begin{tabular}{lcccccccc}
\toprule
Category & I-AUROC & I-AP & I-F1 & P-AUROC & P-AP & P-F1 & AUPRO & PRO \\
\midrule
candle & 98.45 & 98.48 & 93.81 & 99.52 & 46.78 & 48.56 & 98.44 & 95.13 \\
capsules & 99.43 & 99.63 & 99.01 & 99.63 & 68.74 & 66.35 & 98.76 & 90.88 \\
cashew & 99.58 & 99.81 & 98.49 & 98.41 & 68.80 & 67.29 & 95.26 & 82.57 \\
chewinggum & 99.34 & 99.72 & 98.99 & 99.28 & 73.85 & 71.91 & 97.94 & 81.78 \\
fryum & 99.46 & 99.76 & 98.02 & 96.21 & 46.35 & 51.86 & 87.66 & 87.49 \\
macaroni1 & 99.47 & 99.52 & 96.59 & 99.71 & 27.36 & 34.25 & 99.06 & 93.12 \\
macaroni2 & 97.71 & 97.70 & 94.12 & 99.82 & 21.09 & 30.69 & 99.44 & 93.06 \\
pcb1 & 98.81 & 98.86 & 95.96 & 99.75 & 87.44 & 79.25 & 99.24 & 91.96 \\
pcb2 & 98.41 & 98.38 & 97.00 & 99.18 & 34.86 & 44.10 & 97.29 & 86.53 \\
pcb3 & 99.73 & 99.75 & 97.98 & 99.30 & 39.27 & 43.38 & 97.67 & 86.72 \\
pcb4 & 100.00 & 100.00 & 100.00 & 98.93 & 52.36 & 52.82 & 96.45 & 90.83 \\
pipe\_fryum & 99.86 & 99.93 & 99.50 & 99.12 & 60.40 & 66.14 & 97.14 & 94.96 \\
\textbf{Mean} & \textbf{99.19} & \textbf{99.30} & \textbf{97.46} & \textbf{99.07} & \textbf{52.28} & \textbf{54.72} & \textbf{97.03} & \textbf{89.59} \\
\bottomrule
\end{tabular}
\caption{VisA per-category results for ProCon at the final $5\%$ coreset setting.}
\label{tab:s_visa_percat}
\end{table*}

\subsection{Real-IAD Final Results at $1\%$ Coreset}

Due to the larger number of categories, Real-IAD results are split into two tables.

\begin{table*}[t]
\centering
\begin{tabular}{lcccccccc}
\toprule
Category & I-AUROC & I-AP & I-F1 & P-AUROC & P-AP & P-F1 & AUPRO & PRO \\
\midrule
audiojack & 94.73 & 91.70 & 83.49 & 99.65 & 54.29 & 55.81 & 98.85 & 93.81 \\
bottle\_cap & 96.75 & 96.15 & 89.22 & 99.81 & 41.29 & 41.90 & 99.38 & 95.57 \\
button\_battery & 90.79 & 91.61 & 86.14 & 99.32 & 48.37 & 57.59 & 97.79 & 90.27 \\
end\_cap & 92.57 & 92.90 & 87.82 & 99.46 & 27.57 & 35.61 & 98.25 & 94.00 \\
eraser & 95.65 & 94.58 & 85.37 & 99.73 & 47.76 & 49.60 & 99.14 & 88.11 \\
fire\_hood & 89.27 & 83.24 & 73.49 & 99.56 & 41.93 & 46.83 & 98.54 & 91.20 \\
mint & 84.82 & 84.75 & 75.76 & 98.45 & 28.96 & 38.22 & 95.56 & 86.54 \\
mounts & 88.89 & 79.22 & 77.61 & 99.34 & 45.09 & 46.56 & 97.90 & 92.85 \\
pcb & 95.08 & 97.01 & 90.27 & 99.66 & 61.85 & 61.41 & 98.90 & 96.38 \\
phone\_battery & 94.38 & 92.41 & 84.75 & 99.76 & 70.02 & 63.97 & 99.22 & 95.05 \\
plastic\_nut & 95.42 & 91.58 & 84.68 & 99.85 & 53.29 & 51.61 & 99.53 & 97.83 \\
plastic\_plug & 92.69 & 90.12 & 80.07 & 99.28 & 34.59 & 39.33 & 97.71 & 91.16 \\
porcelain\_doll & 86.51 & 75.98 & 70.91 & 98.96 & 31.80 & 37.26 & 96.58 & 93.60 \\
regulator & 91.67 & 85.55 & 75.91 & 99.68 & 51.91 & 54.50 & 99.00 & 92.32 \\
rolled\_strip\_base & 99.53 & 99.75 & 98.63 & 99.76 & 37.13 & 45.09 & 99.21 & 92.56 \\
\bottomrule
\end{tabular}
\caption{Real-IAD per-category results for ProCon at the final $1\%$ coreset setting, part 1.}
\label{tab:s_realiad_percat_1}
\end{table*}

\begin{table*}[t]
\centering
\begin{tabular}{lcccccccc}
\toprule
Category & I-AUROC & I-AP & I-F1 & P-AUROC & P-AP & P-F1 & AUPRO & PRO \\
\midrule
sim\_card\_set & 97.71 & 97.99 & 92.18 & 99.49 & 67.70 & 63.42 & 98.57 & 91.32 \\
switch & 98.96 & 99.15 & 95.46 & 97.43 & 68.58 & 65.76 & 93.77 & 94.17 \\
tape & 98.92 & 98.28 & 93.78 & 99.86 & 62.64 & 59.40 & 99.54 & 96.02 \\
terminalblock & 98.84 & 99.12 & 95.85 & 99.86 & 58.22 & 55.35 & 99.54 & 94.36 \\
toothbrush & 87.88 & 88.79 & 82.81 & 96.81 & 34.70 & 41.06 & 89.92 & 86.98 \\
toy & 89.82 & 91.21 & 86.45 & 93.19 & 24.56 & 35.48 & 82.19 & 85.02 \\
toy\_brick & 82.36 & 79.18 & 70.21 & 97.49 & 40.86 & 44.85 & 92.63 & 77.80 \\
transistor1 & 98.31 & 98.73 & 94.52 & 99.57 & 59.35 & 56.66 & 98.66 & 91.41 \\
u\_block & 95.15 & 93.07 & 85.15 & 99.68 & 57.09 & 58.70 & 98.96 & 88.98 \\
usb & 95.48 & 94.66 & 87.79 & 99.29 & 50.10 & 51.60 & 97.91 & 92.04 \\
usb\_adaptor & 86.03 & 79.57 & 73.50 & 99.40 & 32.76 & 37.76 & 98.22 & 85.61 \\
vcpill & 95.34 & 94.50 & 87.01 & 99.00 & 68.52 & 67.42 & 96.89 & 88.73 \\
wooden\_beads & 92.75 & 92.11 & 84.25 & 99.33 & 56.44 & 56.91 & 97.95 & 91.44 \\
woodstick & 89.05 & 82.06 & 73.21 & 99.41 & 59.14 & 59.13 & 98.27 & 88.45 \\
zipper & 99.26 & 99.60 & 96.91 & 99.11 & 63.84 & 65.76 & 97.05 & 93.64 \\
\textbf{Mean} & \textbf{93.15} & \textbf{91.15} & \textbf{84.77} & \textbf{99.04} & \textbf{49.35} & \textbf{51.49} & \textbf{97.19} & \textbf{91.24} \\
\bottomrule
\end{tabular}
\caption{Real-IAD per-category results for ProCon at the final $1\%$ coreset setting, part 2.}
\label{tab:s_realiad_percat_2}
\end{table*}

\section{Cross-domain Generalization}
\label{sec:s_crossdomain}

To probe whether the final ProCon recipe transfers beyond the MVTec-AD, VisA, and Real-IAD consumer-object family, we run the unchanged recipe (the final four-layer pool $\{-3,-6,-8,-9\}$, $1\%$ coreset budget, $B=1$, no re-tuning) on three additional benchmarks spanning very different domains: MPDD~\cite{jezek2021mpdd}, a metal-part defect benchmark with $6$ categories; BTAD~\cite{mishra2021vtadl}, an industrial texture and product benchmark with $3$ categories; and Uni-Medical, the medical anomaly detection suite from BMAD~\cite{bao2024bmad}, using its $3$ pixel-mask subsets brain (BraTS2021), liver, and retina-RESC. We use $B=1$ throughout, since the bank-count sweep (Table~\ref{tab:s_banks}) established that a single per-layer coreset already captures essentially all of the localization signal; this keeps the transfer test cheap while changing no design choice.

Table~\ref{tab:s_crossdomain_summary} reports category-averaged results across all eight metrics at seed $0$. Localization transfers across all three domains without any re-tuning: P-AUROC is at least $0.972$ and AUPRO at least $0.908$ on every dataset, matching the strong localization the final recipe shows on the main benchmarks. This is direct evidence that the layer pool $\{-3,-6,-8,-9\}$ is a domain-agnostic choice rather than one fit to consumer-object textures.

\begin{table}[t]
\centering
\setlength{\tabcolsep}{3pt}
\begin{tabular}{lcccc}
\toprule
Dataset & I-AUROC & P-AUROC & P-AP & AUPRO \\
\midrule
MPDD ($6$) & 0.9740 & 0.9786 & 0.5277 & 0.9359 \\
BTAD ($3$) & 0.9515 & 0.9778 & 0.7137 & 0.9292 \\
Uni-Medical ($3$) & 0.8767 & 0.9716 & 0.5594 & 0.9075 \\
\bottomrule
\end{tabular}
\caption{Cross-domain summary, category-averaged, seed $0$, $B=1$, unchanged final recipe. Full eight-metric per-category results are in Tables~\ref{tab:s_mpdd}--\ref{tab:s_unimedical}.}
\label{tab:s_crossdomain_summary}
\end{table}

Three findings hold across the per-category tables below. First, MPDD image-level detection is near-saturated (mean I-AUROC $0.974$; connector, metal\_plate, and bracket\_white reach $1.000$), so the final recipe is competitive on a metal-part benchmark it was never tuned on. Second, the apparent P-AP spread is a defect-size artifact rather than a failure: within MPDD the large-defect classes (metal\_plate $0.972$, tubes $0.829$, connector $0.761$ P-AP) score high while the tiny-defect brackets (bracket\_white $0.079$) collapse P-AP by construction, exactly the low-pixel-fraction ceiling seen on Real-IAD, while their P-AUROC and AUPRO stay above $0.99$ and $0.96$. Third, the one genuinely hard case is liver histopathology at I-AUROC $0.757$: distinguishing normal from abnormal tissue texture is where a frozen natural-image DINOv2 backbone is weakest, yet even there localization holds (P-AUROC $0.975$).

\begin{table}[t]
\centering
\setlength{\tabcolsep}{3pt}
\begin{tabular}{lcccccccc}
\toprule
Category & I-AUROC & I-AP & I-F1 & P-AUROC & P-AP & P-F1 & AUPRO & PRO \\
\midrule
bracket\_black & 0.9089 & 0.9542 & 0.8889 & 0.9875 & 0.2884 & 0.3981 & 0.9603 & 0.9338 \\
bracket\_brown & 0.9646 & 0.9767 & 0.9714 & 0.9155 & 0.2369 & 0.3174 & 0.7542 & 0.8437 \\
bracket\_white & 1.0000 & 1.0000 & 1.0000 & 0.9963 & 0.0786 & 0.1623 & 0.9895 & 0.9480 \\
connector & 1.0000 & 1.0000 & 1.0000 & 0.9872 & 0.7611 & 0.7098 & 0.9596 & 0.9099 \\
metal\_plate & 1.0000 & 1.0000 & 1.0000 & 0.9940 & 0.9717 & 0.9136 & 0.9805 & 0.9633 \\
tubes & 0.9706 & 0.9884 & 0.9552 & 0.9911 & 0.8294 & 0.7706 & 0.9714 & 0.9130 \\
\textbf{Mean} & \textbf{0.9740} & \textbf{0.9866} & \textbf{0.9693} & \textbf{0.9786} & \textbf{0.5277} & \textbf{0.5453} & \textbf{0.9359} & \textbf{0.9186} \\
\bottomrule
\end{tabular}
\caption{MPDD per-category results, unchanged final recipe, $B=1$, seed $0$.}
\label{tab:s_mpdd}
\end{table}

\begin{table}[t]
\centering
\setlength{\tabcolsep}{3pt}
\begin{tabular}{lcccccccc}
\toprule
Category & I-AUROC & I-AP & I-F1 & P-AUROC & P-AP & P-F1 & AUPRO & PRO \\
\midrule
01 & 0.9815 & 0.9941 & 0.9796 & 0.9637 & 0.5597 & 0.5722 & 0.8836 & 0.6991 \\
02 & 0.8757 & 0.9801 & 0.9496 & 0.9714 & 0.7745 & 0.6992 & 0.9091 & 0.5831 \\
03 & 0.9974 & 0.9642 & 0.9524 & 0.9985 & 0.8070 & 0.7420 & 0.9950 & 0.9235 \\
\textbf{Mean} & \textbf{0.9515} & \textbf{0.9795} & \textbf{0.9605} & \textbf{0.9778} & \textbf{0.7137} & \textbf{0.6711} & \textbf{0.9292} & \textbf{0.7352} \\
\bottomrule
\end{tabular}
\caption{BTAD per-category results, unchanged final recipe, $B=1$, seed $0$.}
\label{tab:s_btad}
\end{table}

\begin{table}[t]
\centering
\setlength{\tabcolsep}{3pt}
\begin{tabular}{lcccccccc}
\toprule
Subset & I-AUROC & I-AP & I-F1 & P-AUROC & P-AP & P-F1 & AUPRO & PRO \\
\midrule
brain & 0.9401 & 0.9867 & 0.9406 & 0.9807 & 0.7073 & 0.6868 & 0.9359 & 0.8075 \\
liver & 0.7567 & 0.6791 & 0.6827 & 0.9745 & 0.2471 & 0.3274 & 0.9149 & 0.8999 \\
retina-RESC & 0.9335 & 0.9290 & 0.8451 & 0.9596 & 0.7239 & 0.6518 & 0.8717 & 0.7793 \\
\textbf{Mean} & \textbf{0.8767} & \textbf{0.8649} & \textbf{0.8228} & \textbf{0.9716} & \textbf{0.5594} & \textbf{0.5553} & \textbf{0.9075} & \textbf{0.8289} \\
\bottomrule
\end{tabular}
\caption{Uni-Medical (BMAD pixel-mask subsets) per-subset results, unchanged final recipe, $B=1$, seed $0$.}
\label{tab:s_unimedical}
\end{table}

\section{Negative Results and Design Lessons}

We report negative results because they clarify why the final ProCon design is not a naive layer ensemble.

\paragraph{Shallow layers can be toxic.}
Layer $-12$ collapses in single-layer profiling, with much lower image AUROC and localization metrics than other layers. We therefore exclude it from all promoted pools.

\paragraph{Per-image robust normalization fails.}
Per-image robust normalization standardizes the score distribution of each test image. This removes useful anomaly contrast and causes a large performance drop in the fusion sweep.

\paragraph{Independent layer-wise memories outperform concatenated memory.}
The concat single-bank baseline underperforms independent layer-wise memories, both under the Phase-2 protocol (Table~\ref{tab:s_fusion}) and at the final operating point (Table~\ref{tab:s_concat}). This supports the central claim that layer-specific normal geometry should be preserved before residual fusion.

\paragraph{Validation P-AP alone can mislead.}
The pool $\{-8,-9\}$ achieves the highest P-AP on the bottom-3 validation subset, but it does not survive full promotion on MVTec-AD and VisA. This motivates the full-promotion decision rule.

\paragraph{Density-biased memory pruning failed.}
A density-biased coreset pruning variant favored dense anchors and reduced memory diversity, causing failures on high-variance categories. This variant was retired and is not used in ProCon.

\section{Additional Qualitative Results}

We provide additional qualitative anomaly maps for MVTec-AD, VisA, and Real-IAD. The main paper keeps only a compact qualitative figure; this section contains the larger diagnostic visualization that was removed from the main body to preserve space for comparison tables and discussion.

\begin{figure*}[t]
\centering
\includegraphics[width=0.95\textwidth]{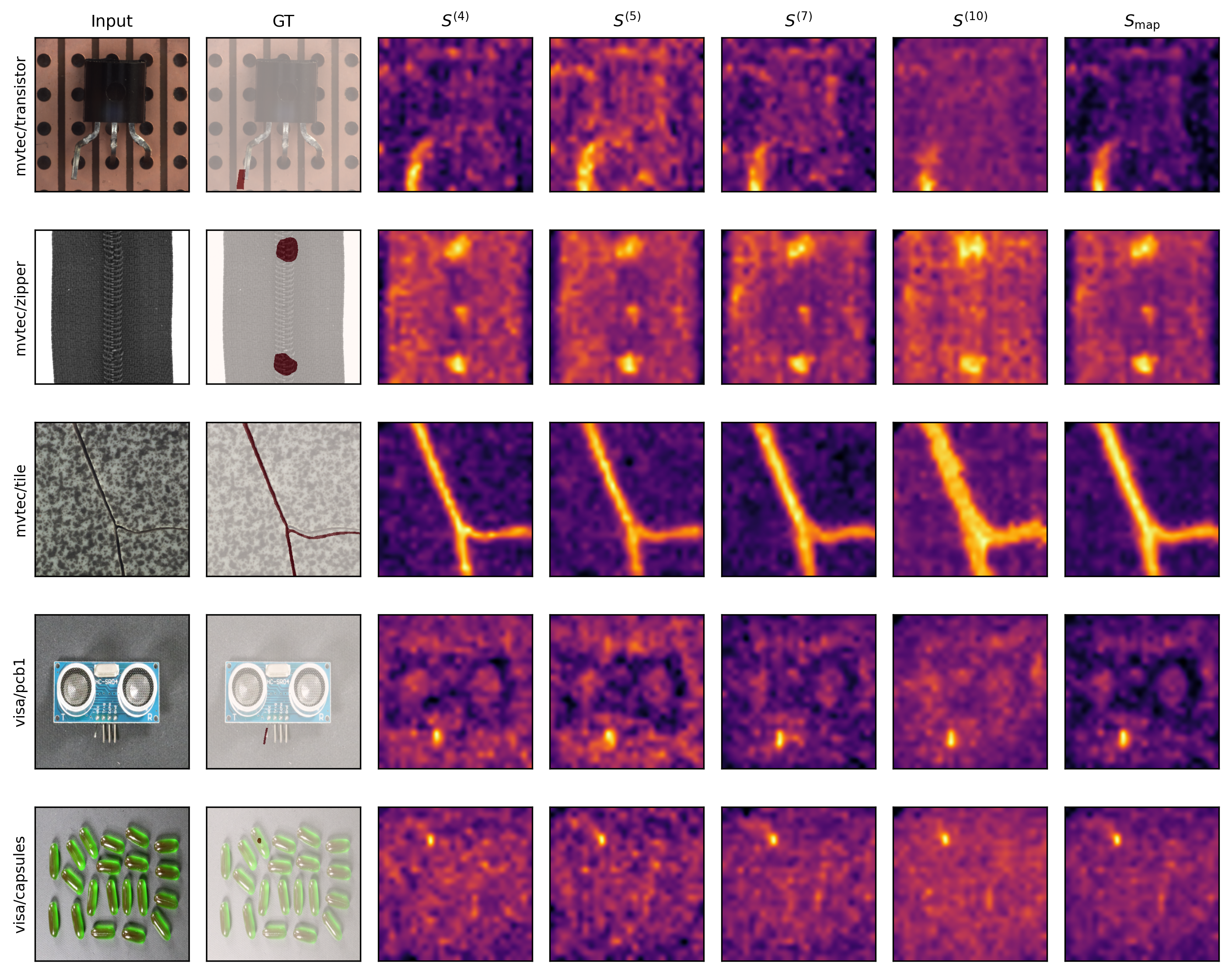}
\caption{Layer-wise residual maps produced by ProCon. Each selected depth generates a residual map under the same soft-projection objective. The final map $S_{\mathrm{map}}$ is obtained by averaging $S_{-3}$, $S_{-6}$, $S_{-8}$, and $S_{-9}$, preserving complementary anomaly evidence while reducing layer-specific noise.}
\label{fig:s_layer_residuals}
\end{figure*}

\clearpage
\section{Additional Qualitative Results}

We provide additional qualitative anomaly maps for MVTec-AD, VisA, and Real-IAD.
Each row compares the input image, ground-truth mask, nearest-neighbor memory,
soft-projection memory, and the final ProCon anomaly map. These visualizations
show that ProCon preserves compact defect responses while reducing the diffuse
background activation often produced by hard nearest-neighbor memory.

\begin{figure*}[p]
\centering
\includegraphics[width=0.92\textwidth]{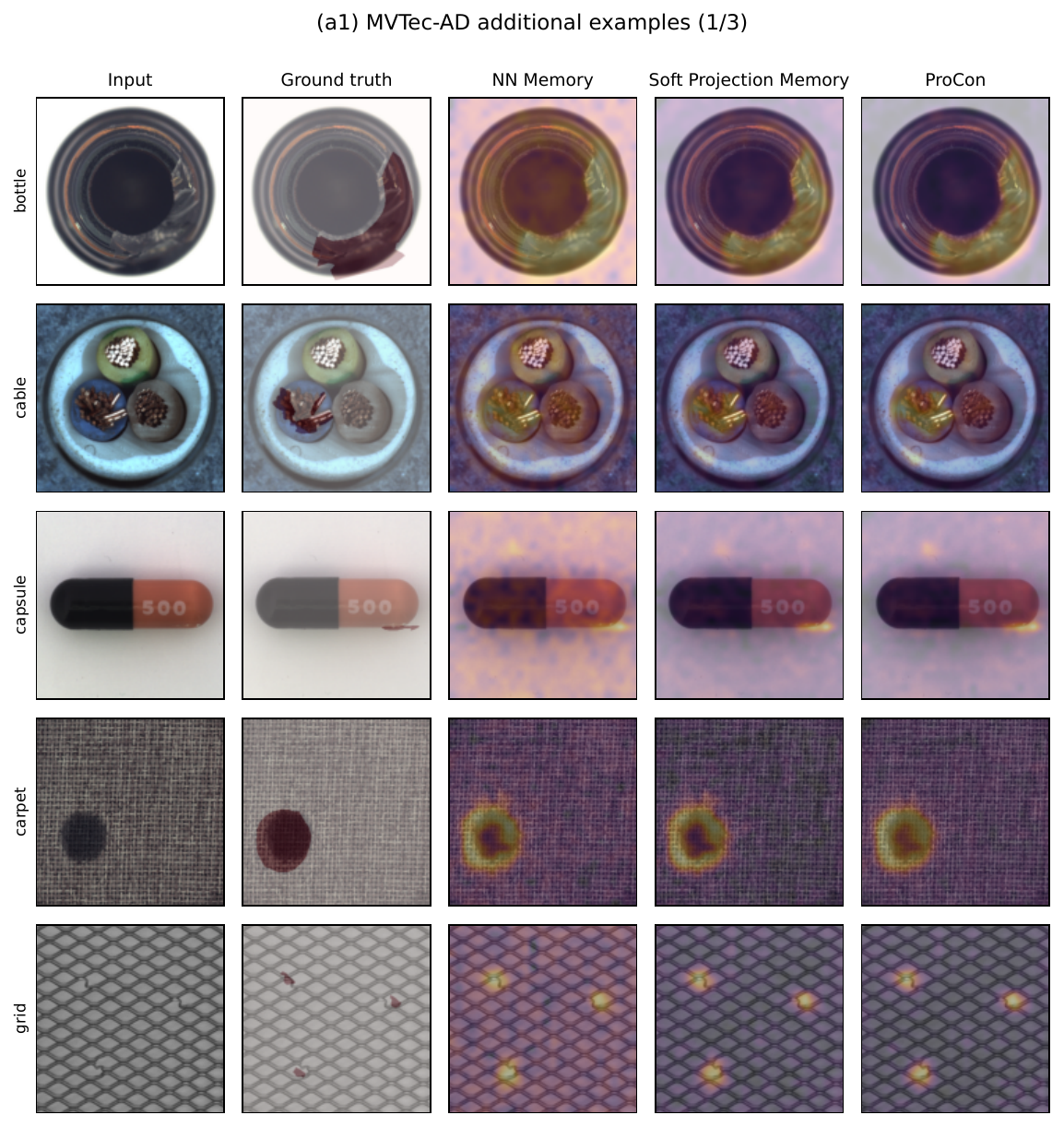}
\caption{Additional qualitative examples on MVTec-AD, part 1. Rows show bottle, cable, capsule, carpet, and grid.}
\label{fig:s_qual_mvtec_1}
\end{figure*}

\begin{figure*}[p]
\centering
\includegraphics[width=0.92\textwidth]{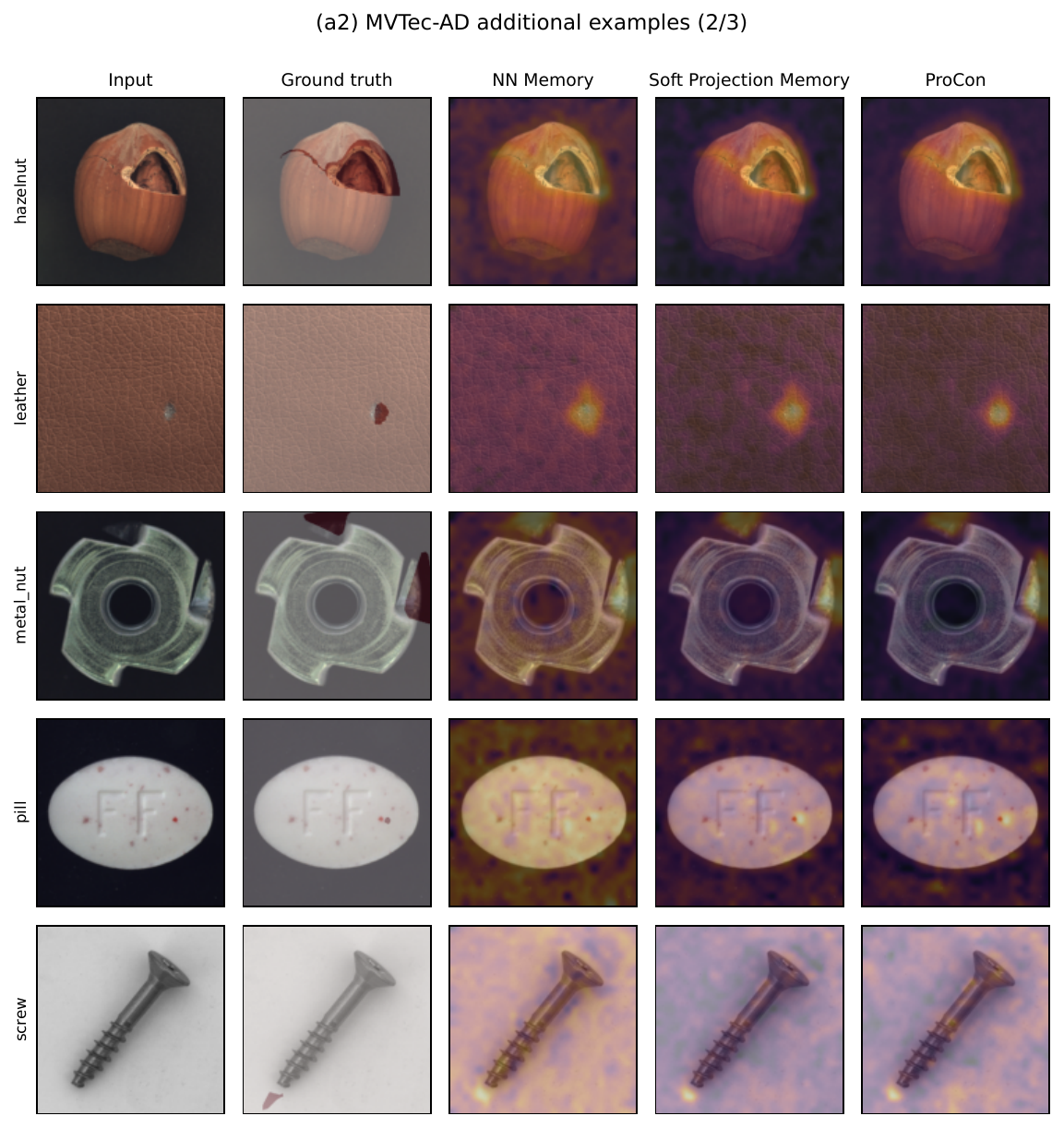}
\caption{Additional qualitative examples on MVTec-AD, part 2. Rows show hazelnut, leather, metal nut, pill, and screw.}
\label{fig:s_qual_mvtec_2}
\end{figure*}

\begin{figure*}[p]
\centering
\includegraphics[width=0.92\textwidth]{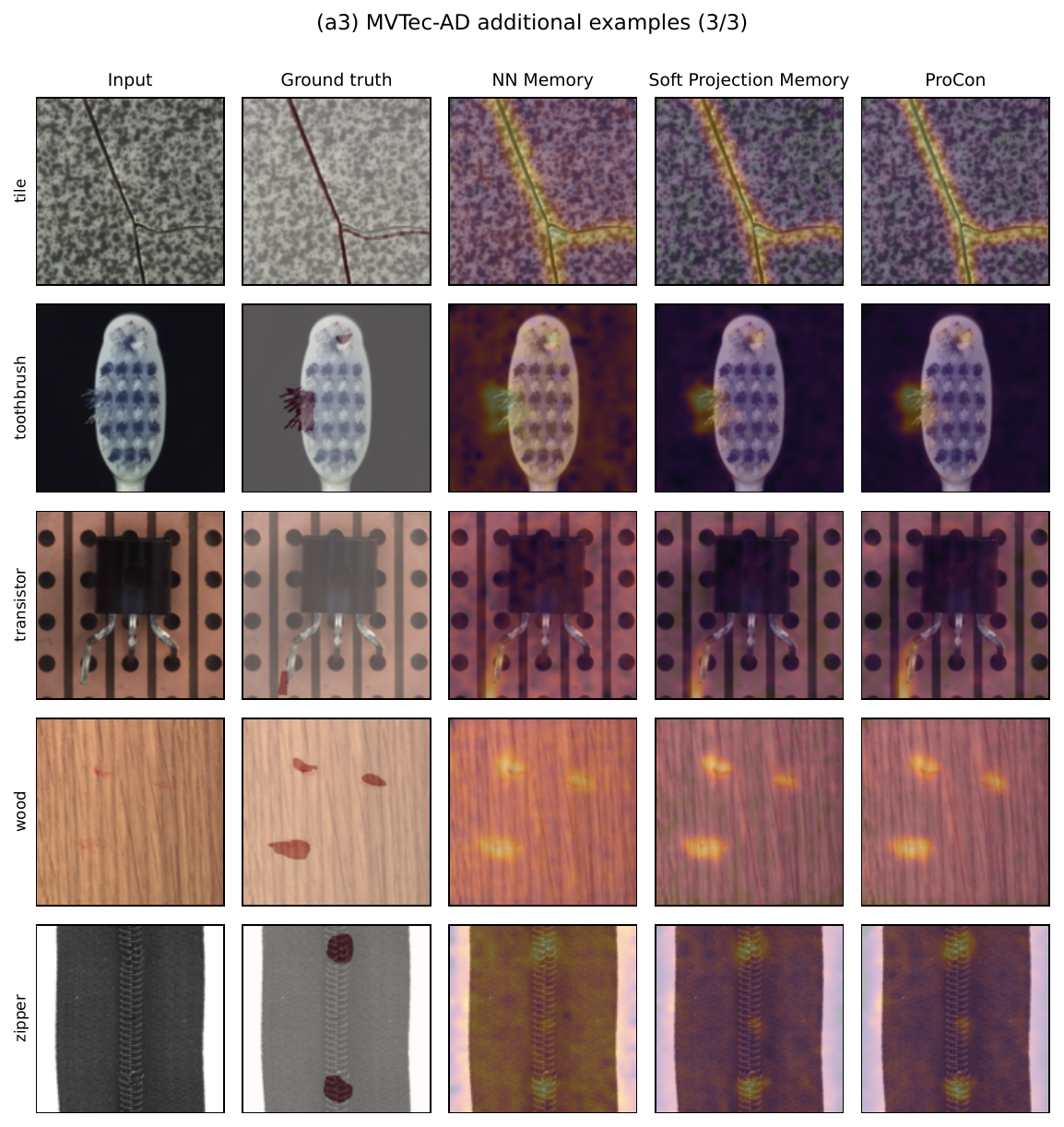}
\caption{Additional qualitative examples on MVTec-AD, part 3. Rows show tile, toothbrush, transistor, wood, and zipper.}
\label{fig:s_qual_mvtec_3}
\end{figure*}

\begin{figure*}[p]
\centering
\includegraphics[width=0.92\textwidth]{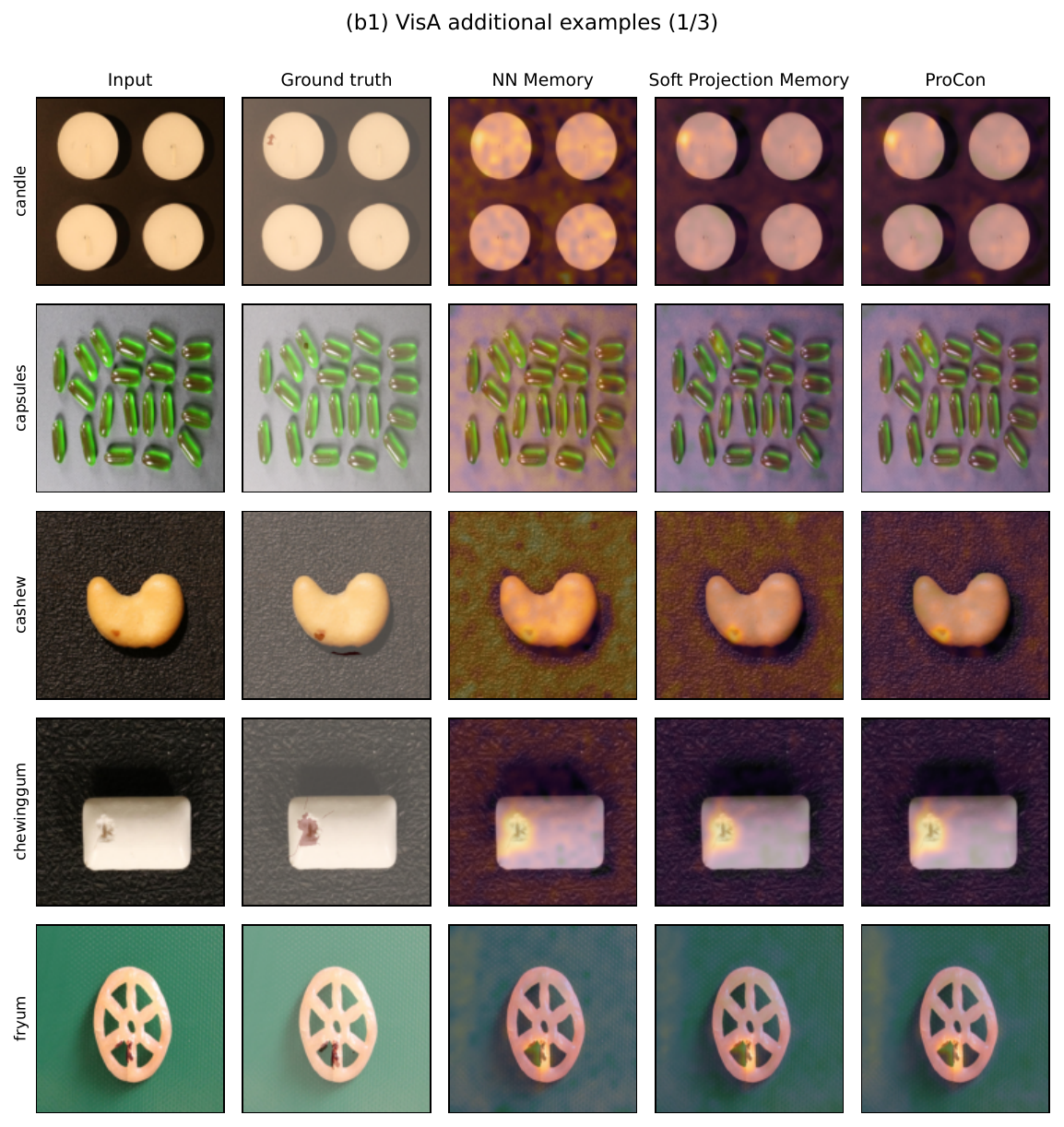}
\caption{Additional qualitative examples on VisA, part 1. Rows show candle, capsules, cashew, chewing gum, and fryum.}
\label{fig:s_qual_visa_1}
\end{figure*}

\begin{figure*}[p]
\centering
\includegraphics[width=0.92\textwidth]{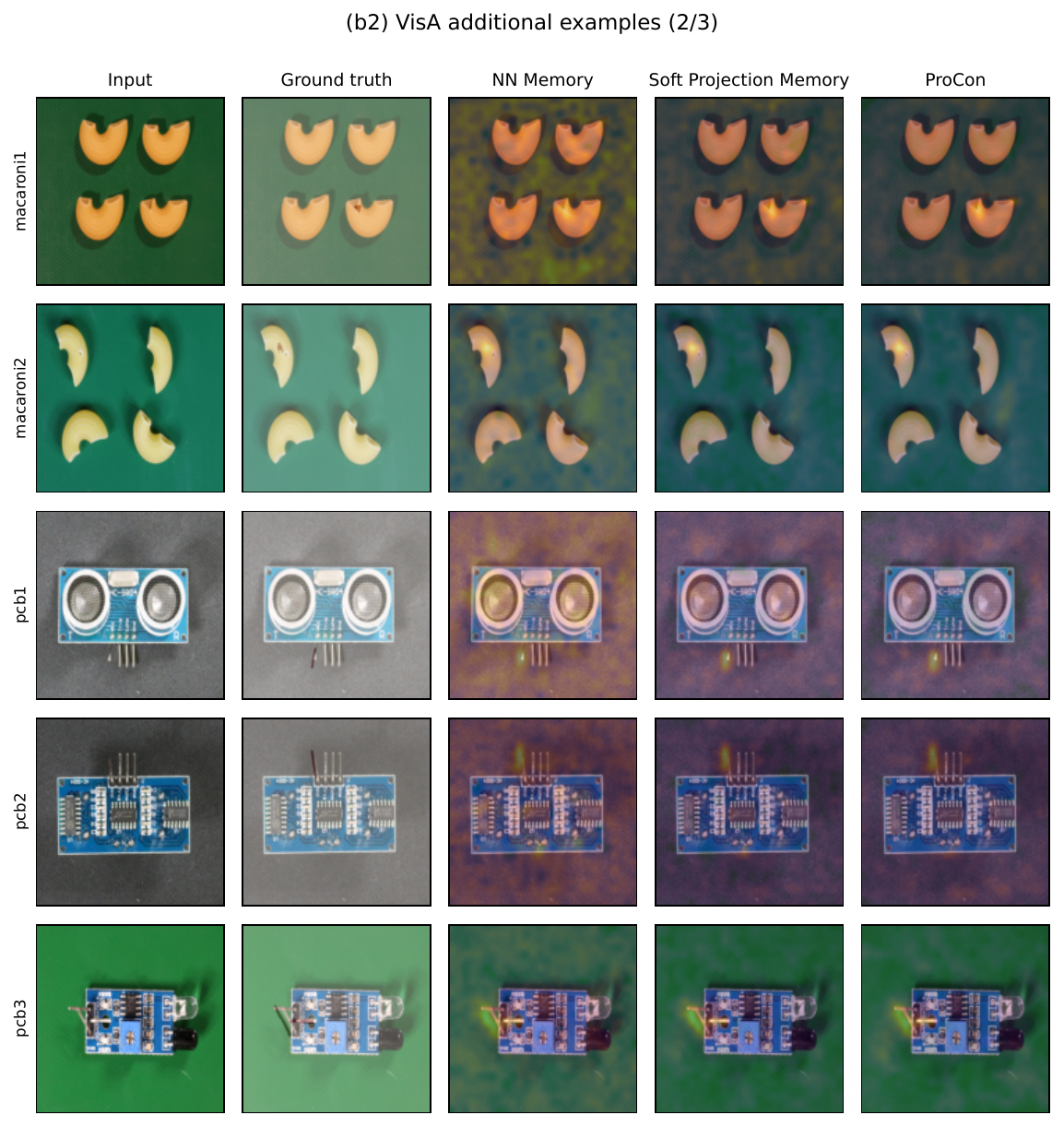}
\caption{Additional qualitative examples on VisA, part 2. Rows show macaroni1, macaroni2, pcb1, pcb2, and pcb3.}
\label{fig:s_qual_visa_2}
\end{figure*}

\begin{figure*}[p]
\centering
\includegraphics[width=0.92\textwidth]{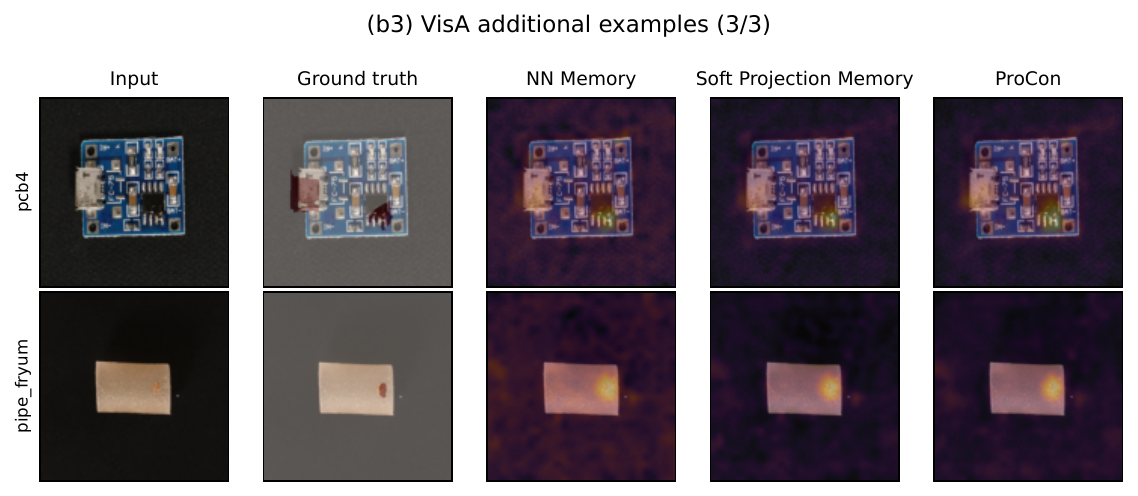}
\caption{Additional qualitative examples on VisA, part 3. Rows show pcb4 and pipe fryum.}
\label{fig:s_qual_visa_3}
\end{figure*}

\begin{figure*}[p]
\centering
\includegraphics[width=0.92\textwidth]{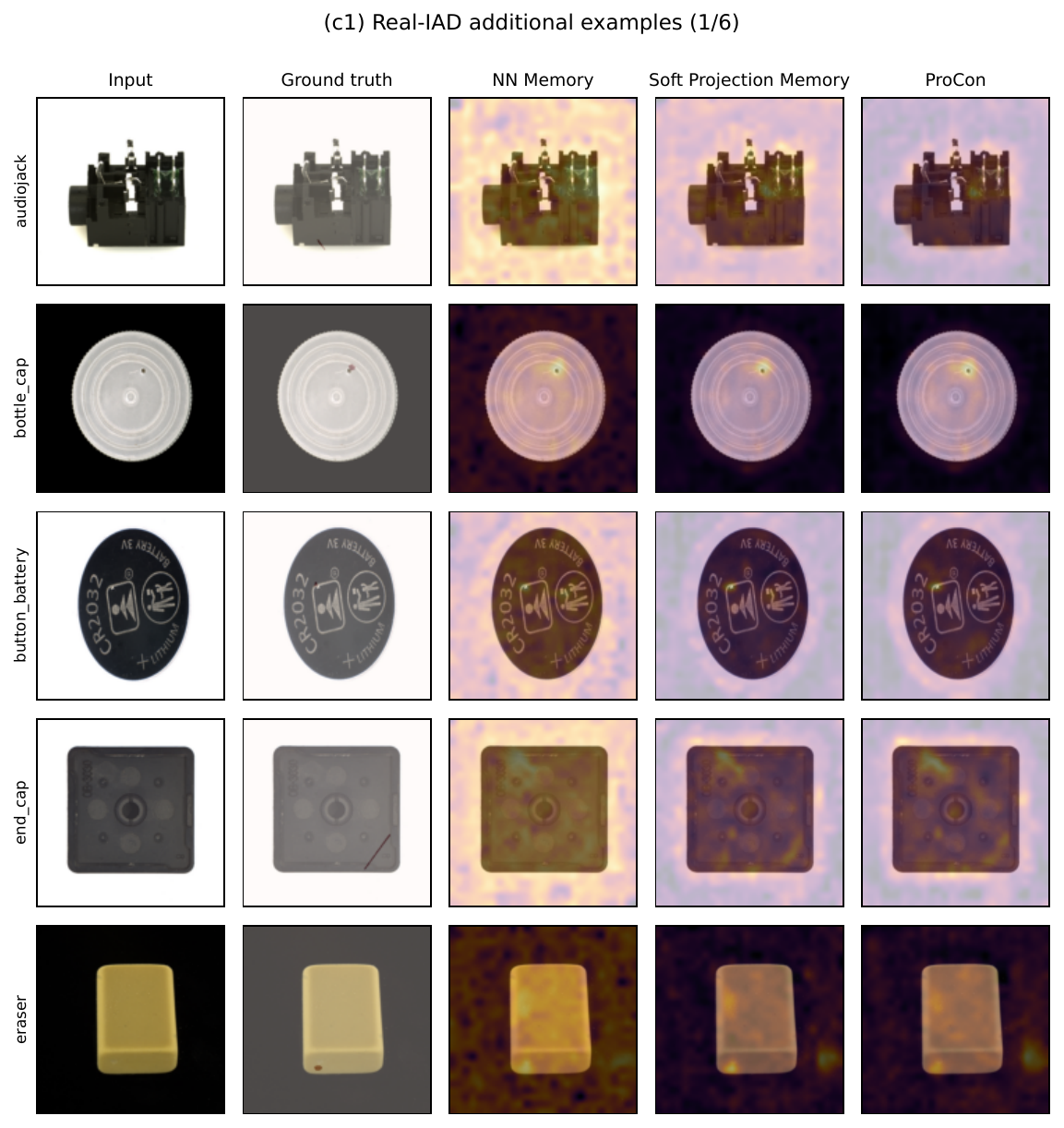}
\caption{Additional qualitative examples on Real-IAD, part 1. Rows show audiojack, bottle cap, button battery, end cap, and eraser.}
\label{fig:s_qual_realiad_1}
\end{figure*}

\begin{figure*}[p]
\centering
\includegraphics[width=0.92\textwidth]{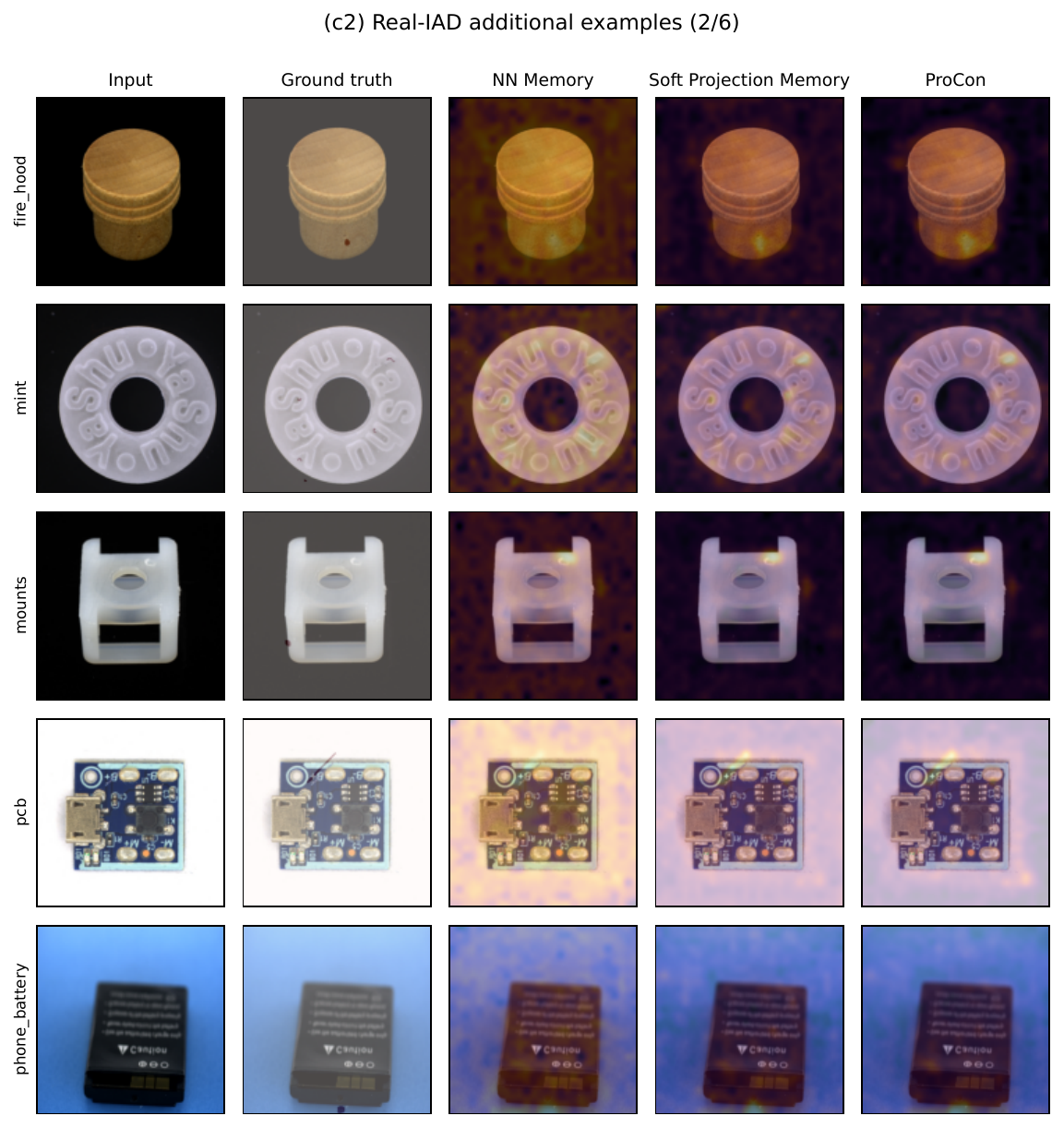}
\caption{Additional qualitative examples on Real-IAD, part 2. Rows show fire hood, mint, mounts, pcb, and phone battery.}
\label{fig:s_qual_realiad_2}
\end{figure*}

\begin{figure*}[p]
\centering
\includegraphics[width=0.92\textwidth]{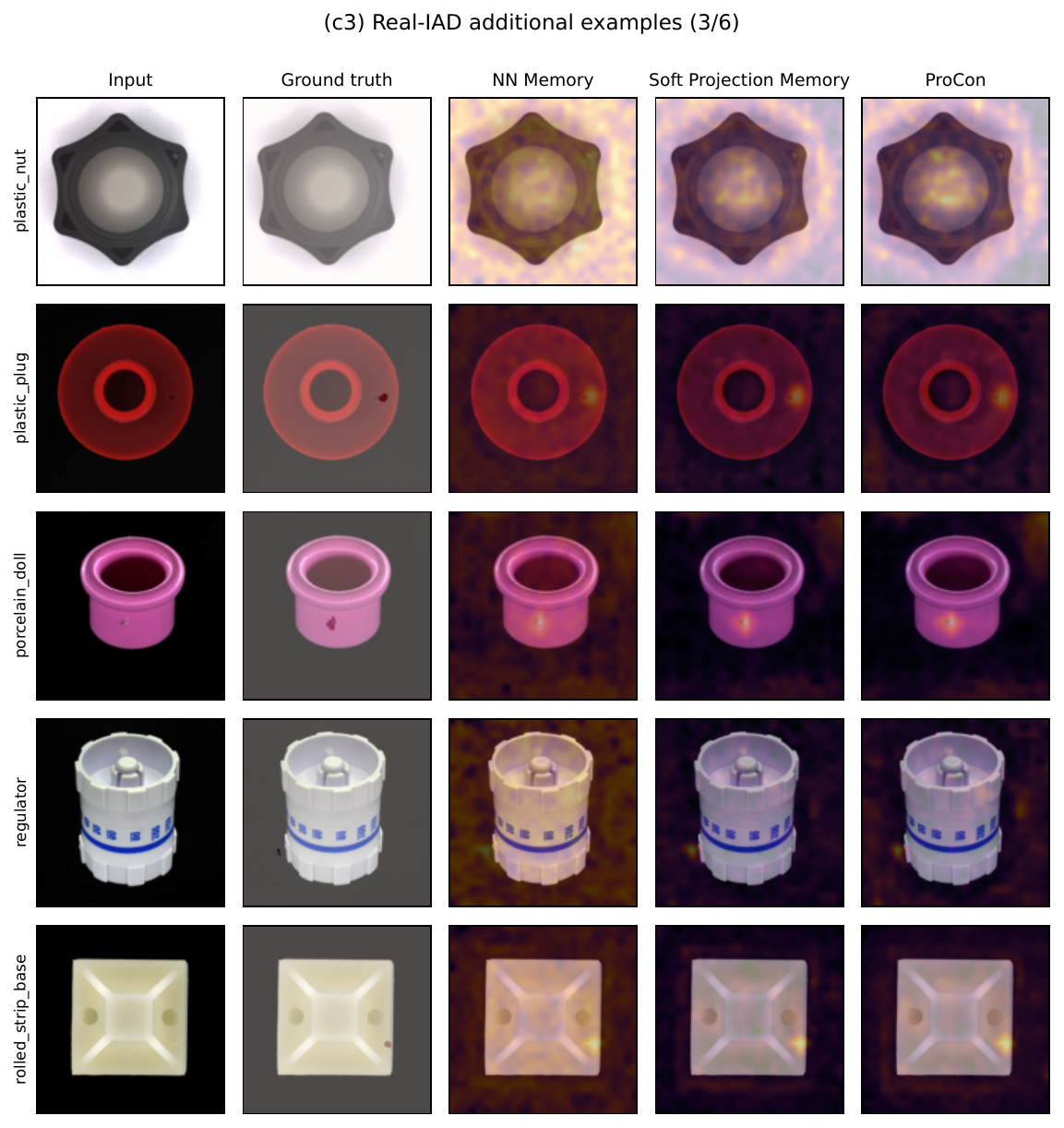}
\caption{Additional qualitative examples on Real-IAD, part 3. Rows show plastic nut, plastic plug, porcelain doll, regulator, and rolled strip base.}
\label{fig:s_qual_realiad_3}
\end{figure*}

\begin{figure*}[p]
\centering
\includegraphics[width=0.92\textwidth]{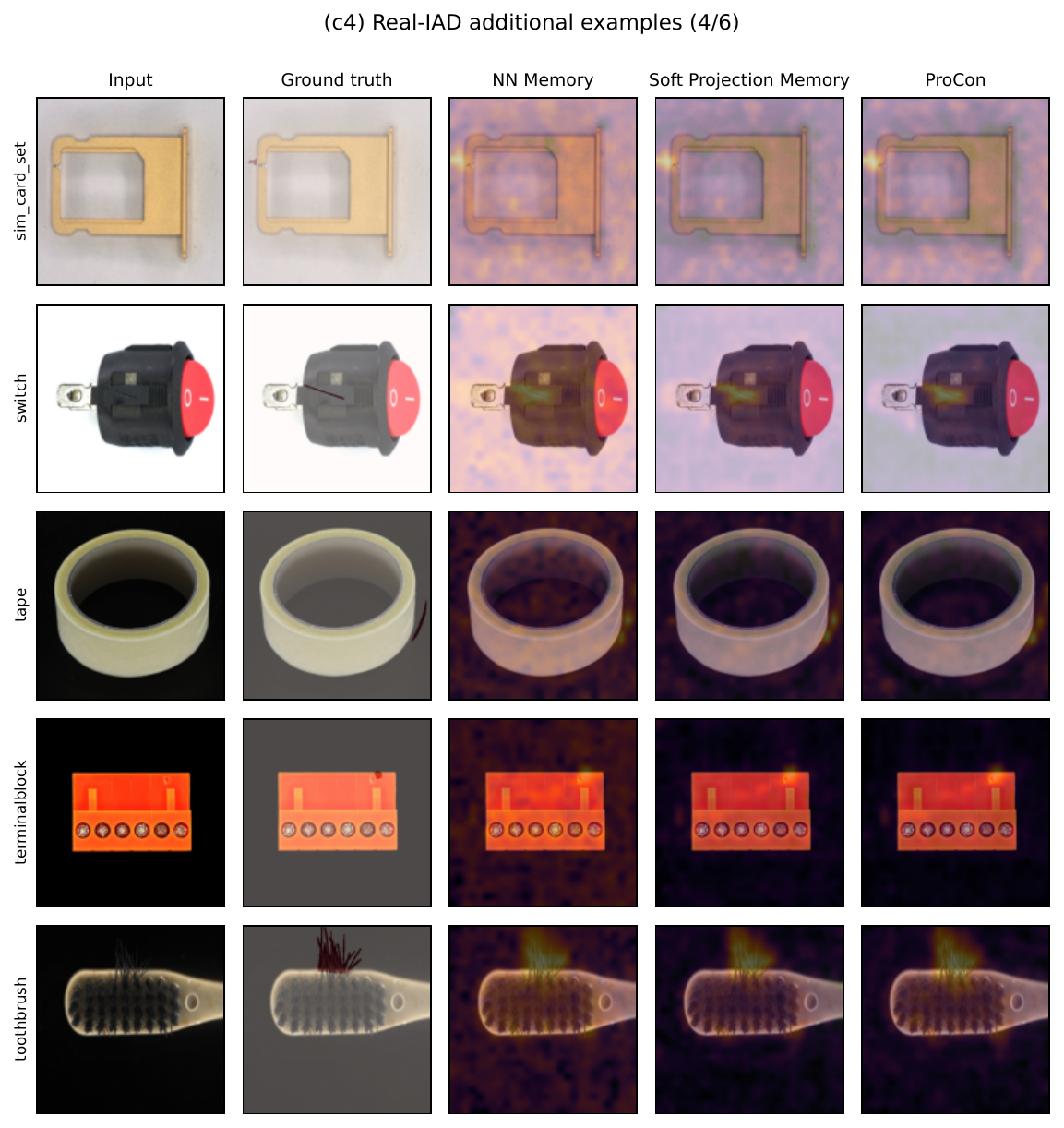}
\caption{Additional qualitative examples on Real-IAD, part 4. Rows show sim card set, switch, tape, terminalblock, and toothbrush.}
\label{fig:s_qual_realiad_4}
\end{figure*}

\begin{figure*}[p]
\centering
\includegraphics[width=0.92\textwidth]{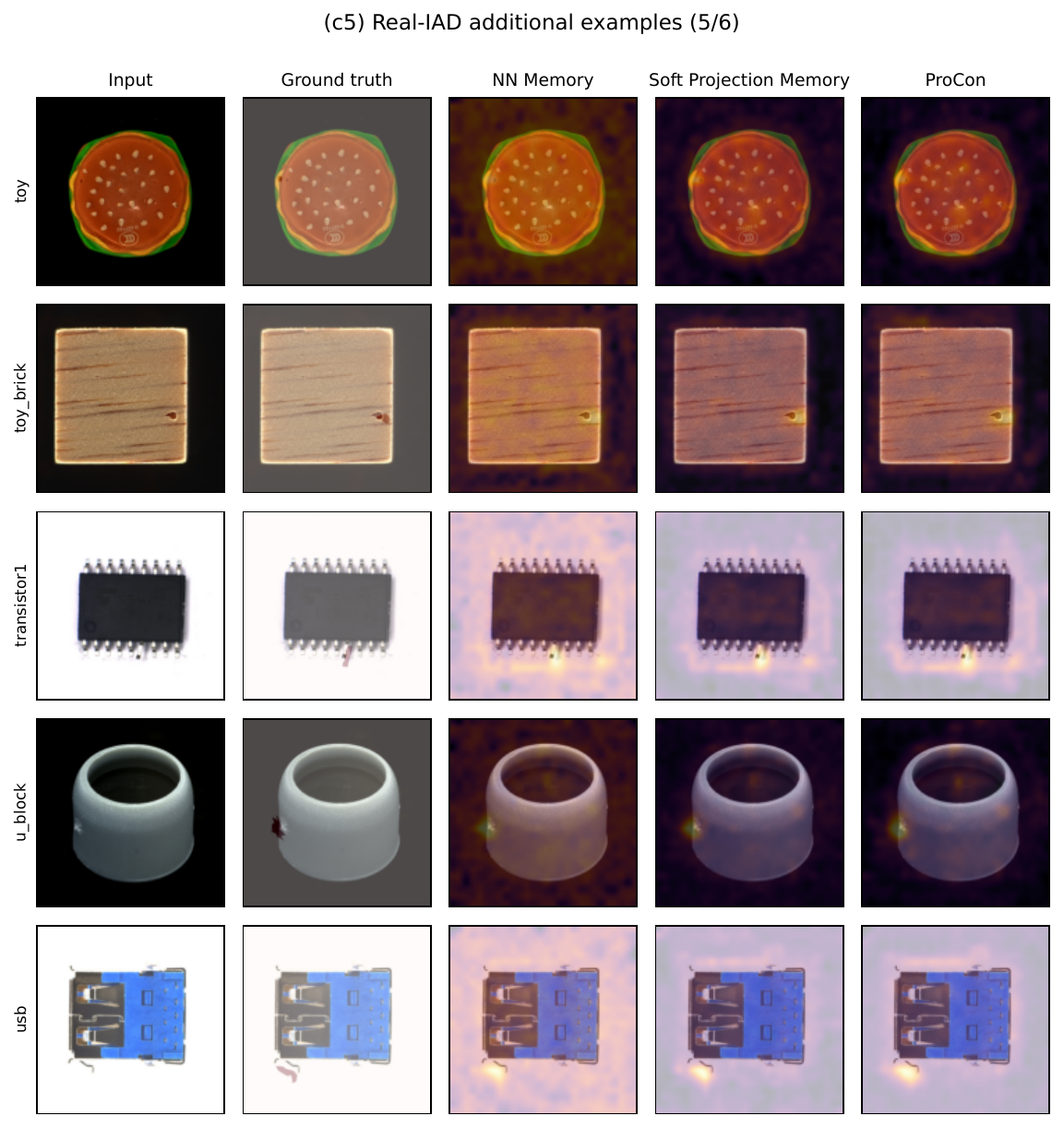}
\caption{Additional qualitative examples on Real-IAD, part 5. Rows show toy, toy brick, transistor1, u block, and usb.}
\label{fig:s_qual_realiad_5}
\end{figure*}

\begin{figure*}[p]
\centering
\includegraphics[width=0.92\textwidth]{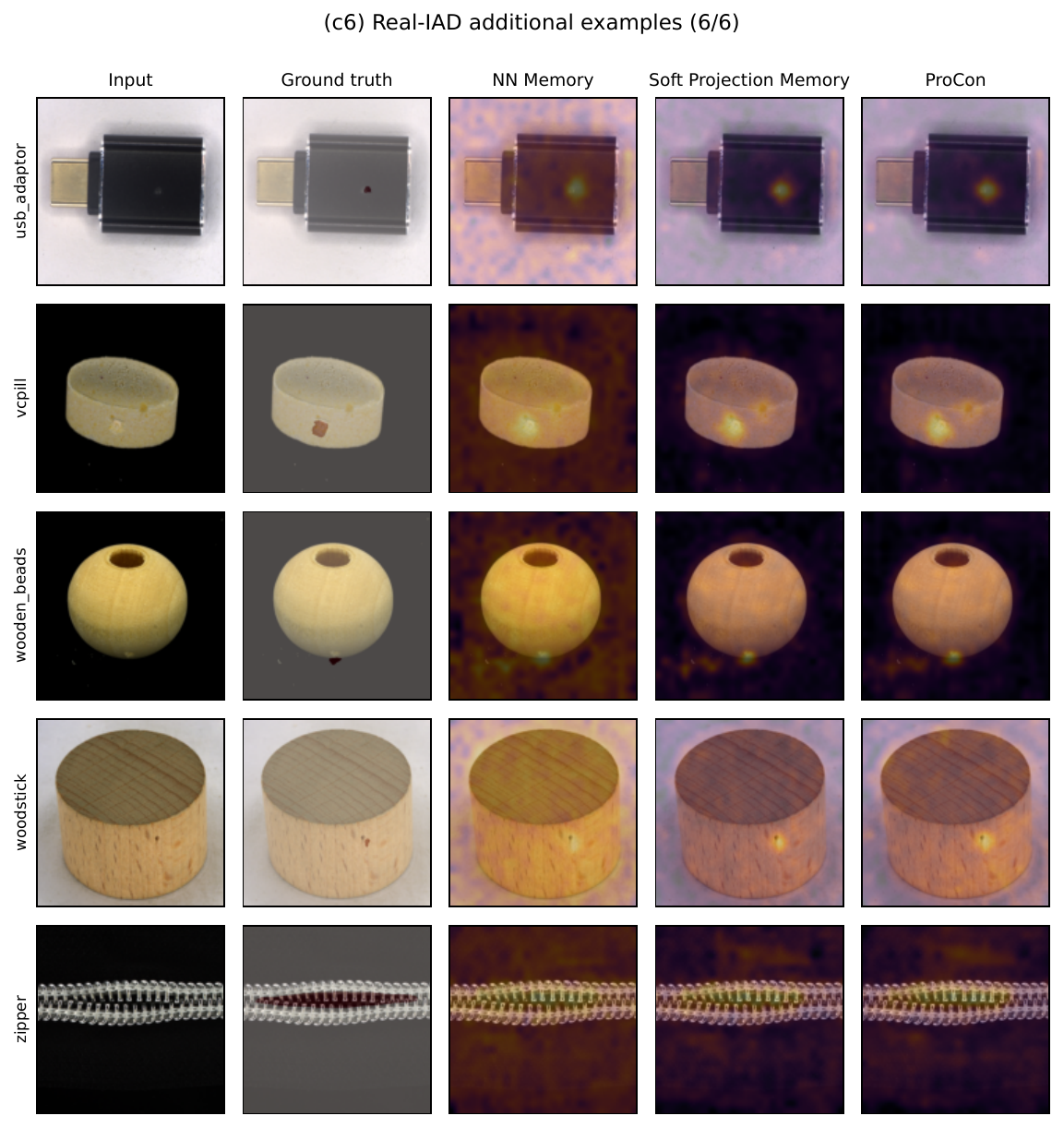}
\caption{Additional qualitative examples on Real-IAD, part 6. Rows show usb adaptor, vcpill, wooden beads, woodstick, and zipper.}
\label{fig:s_qual_realiad_6}
\end{figure*}

\clearpage
\section{Reproducibility Details}

Table~\ref{tab:s_recipe} summarizes the final ProCon recipe. The entire pipeline is training-free: no decoder is trained, the DINOv2 backbone is frozen, no fusion weights are learned, and no pseudo anomalies are used. All reported results use a single fixed evaluation seed; the bank construction seeds $b=1,\dots,B$ are the only stochastic component and their effect is absorbed by the median consensus.

\begin{table}[t]
\centering
\begin{tabular}{ll}
\toprule
Component & Value \\
\midrule
Backbone & frozen DINOv2 ViT-B/14 \\
Input size & $392 \times 392$ \\
Patch grid & $28 \times 28$ \\
Layer pool & $\{-3,-6,-8,-9\}$ \\
Banks per layer & $B=5$ \\
Soft projection neighbors & $k=5$ \\
Temperature & auto, median local distance \\
Bank aggregation & median \\
Layer aggregation & mean \\
Normalization & none \\
Image readout & top-mean, ratio $0.005$ \\
Training & none, coreset memory only \\
\bottomrule
\end{tabular}
\caption{Final ProCon recipe.}
\label{tab:s_recipe}
\end{table}

\begin{table}[t]
\centering
\begin{tabular}{lc}
\toprule
Benchmark & Final Coreset Ratio \\
\midrule
MVTec-AD & $5\%$ \\
VisA & $5\%$ \\
Real-IAD & $1\%$ \\
\bottomrule
\end{tabular}
\caption{Benchmark-specific coreset ratios used for final evaluation.}
\label{tab:s_budget_setting}
\end{table}

\end{document}